\documentclass[accepted]{uai2021} 

\usepackage[american]{babel}

\usepackage{natbib} 
\usepackage{mathtools} 
\usepackage{booktabs} 
\usepackage{tikz} 
\usetikzlibrary{arrows.meta}
\usepackage{floatrow}
\usepackage{bbm}
\usepackage{bm}
\usepackage{algorithm}
\usepackage[noend]{algpseudocode}

\DeclareMathOperator*{\argmax}{argmax}

\title{pRSL: Interpretable Multi--label Stacking by Learning Probabilistic Rules}

\author[1]{\href{mailto:Michael Kirchhof <michael[insert dot]kirchhof[insert at sign]udo[insert dot]edu>?Subject=UAI2021 paper pRSL}{Michael~Kirchhof}{}} 
\author[1]{Lena~Schmid}
\author[2]{Christopher~Reining}
\author[2]{Michael~ten~Hompel}
\author[1]{Markus~Pauly}
\affil[1]{%
    Department of Statistics\\
    TU Dortmund University\\
    Dortmund, Germany
}
\affil[2]{%
    Chair of Material Handling and Warehousing\\
    TU Dortmund University\\
    Dortmund, Germany
}

\begin{document}
\maketitle
\global\csname @topnum\endcsname 0
\global\csname @botnum\endcsname 0

\begin{abstract}
A key task in multi--label classification is modeling the structure between the involved classes. Modeling this structure by probabilistic and interpretable means enables application in a broad variety of tasks such as zero--shot learning or learning from incomplete data. In this paper, we present the probabilistic rule stacking learner (pRSL) which uses probabilistic propositional logic rules and belief propagation to combine the predictions of several underlying classifiers. We derive algorithms for exact and approximate inference and learning, and show that pRSL reaches state--of--the--art performance on various benchmark datasets. 

In the process, we introduce a novel multicategorical generalization of the noisy--or gate. Additionally, we report simulation results on the quality of loopy belief propagation algorithms for approximate inference in bipartite noisy--or networks.
\end{abstract}

\section{Introduction} \label{sec:intro}

\begin{figure}
\floatbox[{\capbeside\thisfloatsetup{capbesideposition={right,top},capbesidewidth=0.5\textwidth}}]{figure}[\FBwidth]
{\caption{Human Activity Recognition: The three embedded ML sensors tell us: $P(\text{overhead work}) = 0.5$,\\ $P(\text{standing}) = 0.95$,\\ $P(\text{hands high}) = 0.7$.\\ Given the rules:\\ \textit{if standing and hands high then often overhead work},\\ \textit{if normal work then almost always hands centered or low},\\ \textit{not standing iff legs moving}, what is the new probability $P(\text{overhead work})$ that the worker does overhead work?}\label{fig:intro}}
{\includegraphics[width=0.4\textwidth]{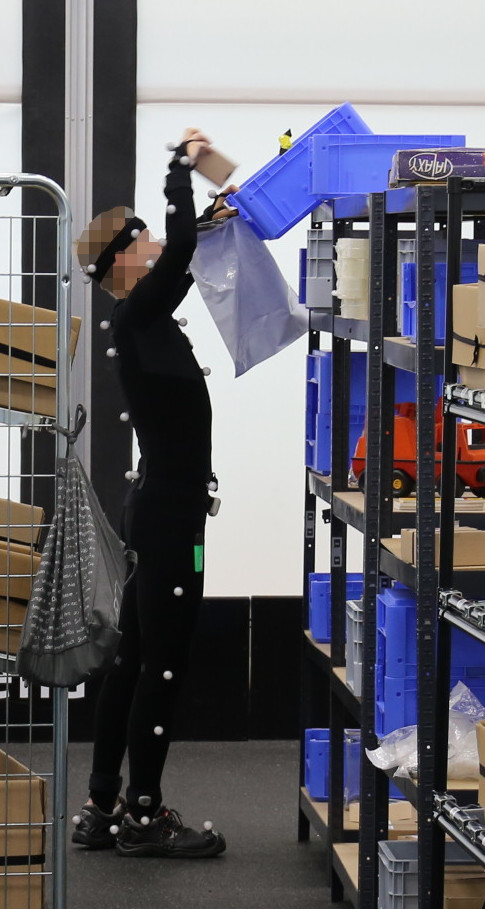}}
\end{figure}

Classifiers that predict a single class have been studied and improved in detail in the past years. However, recent applications in complex real--world environments make it necessary to respect a whole set of different classes and labels \citep{reining2019human} \citep{raies2018silico}. The success of multi--label classifiers has shown that exploiting the structure between those classes is key to exceeding the performance of isolated classifiers \citep{pakrashi2016benchmarking}.

As an example, take the human activity recognition task portrayed in Figure \ref{fig:intro} \citep{niemann2020lara} where we want to detect the actions of a warehouse worker. Three embedded machine learning sensors provide probabilistic information about the worker's stance and hand position and a first guess on whether the worker is performing overhead work. 
Intuitively, we would combine those beliefs to wrap our heads around the situation the worker is in. To decide how likely the worker is working overhead, we may use rules that tell us whether some beliefs contradict or support other beliefs. We can easily formulate such rules like the three exemplary ones given in Figure \ref{fig:intro}. However, the challenging question is how to mathematically apply those non--deterministic rules to the uncertain input beliefs to calculate updated beliefs.

Our model, pRSL, approaches this multi--label classification problem via Bayesian belief propagation along probabilistically generalized propositional logic rules. Thereby, it stays interpretable and allows for a wide variety of relations between the classes. Besides expert--given rules, it can learn rules from data that achieve competitive performance.

This paper is structured as follows: In Section \ref{sec:related}, we discuss existing approaches to multi--label classification and belief combination. In Section \ref{sec:methods}, we define pRSL and explain it along the above example. We propose algorithms for exact and approximate inference, learning rules from possibly incomplete data, and suitable regularization. Section \ref{sec:application} concerns the inference methods' approximation quality and benchmarks of pRSL's against the state--of--the--art. The paper closes with a discussion in Section \ref{sec:conclusion}.

\section{Related Work} \label{sec:related}

This section gives an overview of recent approaches in multi--label learning with an emphasis on models that are stacked on top of ground learners. Such models take the ground learners' probabilistic estimates as input and refine them via modeling the structure between the classes in different ways: We group them into attribute--class, knowledge graph, Bayesian network, and probabilistic rule based methods.

\paragraph{Attribute--Class:} Attribute--class approaches emerged from classical "flat" classification \citep{lampert2013attribute} to give a better insight into the reasoning and utilize knowledge of previously learned classes. They add a layer of semantically interpretable attributes in--between the raw inputs and the output classes. In \citet{atzmon2018LAGO} this layer is connected to the classes via and--or formulas describing each class while grouping similar attributes. \citet{liu2020attribute} model the inter--class structure as a graph where classes sharing similar attributes are close to each other. Attribute--class approaches allow solving zero--shot problems \citep{xian2018zero} in which a new class has to be recognized that has never been shown by example, but only described by its attributes.

\paragraph{Knowledge Graphs:} Other approaches do not split labels into attributes and classes but utilize knowledge graphs instead, thus allowing for broader relations between the labels.
\citet{wu2018multi} use a graph that embodies hierarchical and co--occurrence relations between labels which allows learning even when not all ground truth labels are provided. \citet{lee2018multi} and \citet{liu2020attribute} apply a form of belief propagation where initial beliefs on labels are shared between one another to obtain a joint solution. Knowledge graphs are easy to interpret and applied to large scale problems, but often require ground truth graphs and allow only a few types of heuristic relations between the labels.

\paragraph{Bayesian Networks:} Bayesian networks allow extending these relations to the probabilistic setting. While they have been applied as stand--alone approaches \citep{wang2013randomized}, they experience a recent interest as stacking approaches on top of ground learners \citep{chen2020towards}. For example, \citet{shen2018efui} employ a Bayesian network as an ensemble learner that combines beliefs of several CNNs. A limitation of a Bayesian network is that its directed graph must be acyclic, thus raising design choices that can be difficult outside obvious causal relations.

\paragraph{Probabilistic Rules:} When discussing probabilistic rule learners, a distinction has to be made regarding the goal. Rule mining \citep{mencia2016learning} \citep{pham1995rules} generally evolves around analytical insight, while in multi--label classification rules are learned to improve classification accuracy. Although crisp rules have been explored to define classes in programming APIs \citep{krupka2017toward}, probabilistic rules take uncertainties into account by not completely ruling out contradicting labels, but only weighting them down as in  \citet{ding2015probabilistic}. Moreover, \citet{rapp2020learning} recently proposed a boosting--based method for finding soft if--then rules. Using propositional logic as the starting point for multi--label classification allows formulating complex rules but loses some of the interpretability of the aforementioned graph models.

Our approach borrows principles from all previously described methods: Our model may be stacked onto any ground learners for initial beliefs on the labels or no classifier may be provided to enable zero--shot classification. We use Bayesian networks as underlying framework due to their probabilistic interpretation and the ability to perform fast belief propagation. However, labels are not connected directly to one another but only via rules in a bipartite manner to ensure acyclicity. Rules can take the form of any propositional logic expression and are extended to the probabilistic setting to allow for vast relations between the labels. Like in knowledge graphs, prior knowledge may be incorporated as expert--given rules, but rules can also be learned from data.

\section{Methods} \label{sec:methods}

\subsection{Bayesian Networks} \label{sec:bn}

A Bayesian network \citep{pearl1988} is a compact representation of the joint distribution of $N$ random variables $X_1, \dotsc, X_N$. In our case they are categorical. Each $X_n$ stochastically depends only on a subset of other variables denoted $\text{Pa}(X_n)$ as specified in their conditional probability tables $P(X_n|\text{Pa}(X_n)), n = 1, \dotsc, N$. These dependencies form a directed acyclic graph, where each node is a random variable and each edge shows a dependence. Observations are fed into the network by updating priors or providing hard evidence on the states some variables take. The new information is then propagated through the network to update the beliefs on all variables. In the sequel, we use Bayesian networks as the basis for our novel stacking learner as they feature a transparent structure and probabilistic foundation. 

\subsection{Probabilistic Rule Stacking Learner} \label{sec:prsl}

\begin{figure}
\centering
\begin{tikzpicture}
\node[shape=rectangle,draw=black, minimum size=0.7cm] (x1) at (0,0) {$x_1$};
\node[shape=rectangle,draw=black, minimum size=0.7cm] (x2) at (2,0) {$x_2$};
\node[shape=circle] (x3) at (4,0) {$\ldots$};
\node[shape=rectangle,draw=black, minimum size=0.7cm] (xJ) at (6,0) {$x_J$};

\node[shape=rectangle,draw=black, minimum size=0.7cm] (c1) at (0,1.5) {$C_1$};
\node[shape=rectangle,draw=black, minimum size=0.7cm] (c2) at (2,1.5) {$C_2$};
\node[shape=circle] (c3) at (4,1.5) {$\ldots$};
\node[shape=rectangle,draw=black, minimum size=0.7cm] (cJ) at (6,1.5) {$C_J$};

\node[shape=circle,draw=black] (l1) at (0,3) {$L_1$};
\node[shape=circle,draw=black] (l2) at (2,3) {$L_2$};
\node[shape=circle] (l3) at (4,3) {$\ldots$};
\node[shape=circle,draw=black] (lJ) at (6,3) {$L_J$};

\node[shape=circle,draw=black] (r1) at (1,4.5) {$R_1$};
\node[shape=circle] (r2) at (3,4.5) {$\ldots$};
\node[shape=circle,draw=black] (rK) at (5,4.5) {$R_K$};

\draw[-{Latex[length=2mm]}] (x1) to (c1);
\draw[-{Latex[length=2mm]}] (x2) to (c2);
\draw[-{Latex[length=2mm]}] (xJ) to (cJ);

\draw[-{Latex[length=2mm]}] (c1) to (l1);
\draw[-{Latex[length=2mm]}] (c2) to (l2);
\draw[-{Latex[length=2mm]}] (cJ) to (lJ);

\draw[-{Latex[length=2mm]}] (l1) to (r1);
\draw[-{Latex[length=2mm]}] (l2) to (rK);
\draw[-{Latex[length=2mm]}] (lJ) to (r1);
\draw[-{Latex[length=2mm]}] (lJ) to (rK);
\end{tikzpicture}
\caption{Graphical structure of pRSL. Squares indicate external elements plugged into the framework and circles indicate internal parts of pRSL.}
\label{fig:pRSL}
\end{figure}
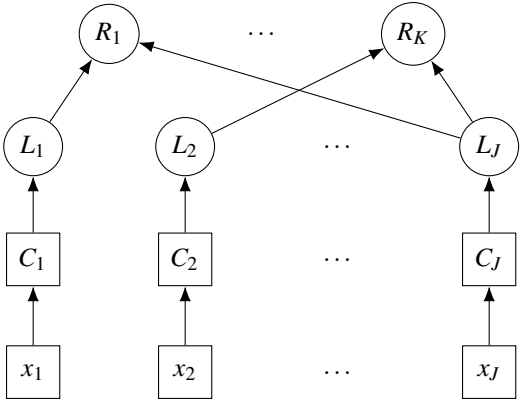

The probabilistic rule stacking learner (pRSL) models the multi--label stacking problem using three kinds of nodes: Classifiers, labels, and rules. The model structure is visualized in Figure \ref{fig:pRSL} and described in the following paragraphs. 

Classifiers $C_j, j = 1, \dotsc, J,$ are plugged into the root of pRSL. Each $C_j$ is an arbitrary machine learning model that accepts some input data $x_j$ and returns a vector of probabilistic estimates $P(C_j | x_j) := (P(C_j = 1 | x_j), \dotsc, P(C_j = M(j) | x_j))$ on its $M(j)$ respective categories\footnote{We use the terms classes and labels interchangeably. They contain an arbitrary number of mutually exclusive categories.} $m = 1, \dotsc, M(j)$. Classifiers are not restricted to predict the same categories or use the same input data. Each classifier is connected to a label node $L_j$. The label nodes contain the same $M(j)$ categories as their connected classifier nodes, but recalibrate the predictions of their classifiers \citep{culakova2020calibrate}. Any function $d_j:[0,1]^{M(j)} \rightarrow [0,1]^{M(j)}$ can be used for calibration, but it can also be the identity in case the classifier already returns predictions that are not over-- or underconfident. The used calibrators are detailed in the experiment sections below. Latent or zero--shot labels are connected to dummy classifiers that always output the labels' priors. As for notation, $\boldsymbol{L} := (L_1, \dotsc, L_J)$ denotes the vector of all labels, $\mathcal{L} := \bigotimes_{j = 1}^J \{1, \dotsc, M(j)\}$ the sample space of all possible categories $\boldsymbol{L}$ can take, and $\boldsymbol{\ell} := (\ell_1, \dotsc , \ell_J) \in \mathcal{L}$ is a particular vector of categories.

The main part of pRSL are the rule nodes $R_k, k = 1, \dotsc, K$. Each rule node may be connected to any selection of label nodes\footnote{To simplify notation, rules depend on all $J$ possible label nodes in the following formulas.} and formulates a propositional logic formula $\varphi_k$ that holds between those labels with a probability $p_k$. The rules' conditional probability table generalizes  truth tables:
\begin{align} \label{form:crisprule}
P(R_k = 1 | \boldsymbol{L} = \boldsymbol{\ell}) = \begin{cases}
p_k&, \boldsymbol{\ell} \models \varphi_k \\
1 - p_k&, \boldsymbol{\ell} \not\models \varphi_k \\
\end{cases},
\end{align}
where $\models$ means that $\boldsymbol{\ell} \in \mathcal{L}$ logically fulfills the formula $\varphi_k$.

To perform inference, first classifiers receive their inputs $\boldsymbol{x} := (x_1, \dotsc, x_J)$. Their estimations $P(C_j | x_j), j = 1, \dotsc, J,$ are then used as priors for the classifier nodes. Then the evidence $\boldsymbol{R} = \boldsymbol{1}$ with $\boldsymbol{R} = (R_1, \dotsc, R_K)$ and $\boldsymbol{1} = (1, \dotsc, 1)$ indicating that all rules are true is handed over to the network. This conditioning opens paths between the labels to communicate and update their priors using belief propagation as described in Section \ref{sec:inference}. Once the propagation has reached an equilibrium state, we obtain the updated beliefs $P(L_j | \boldsymbol{R} = \boldsymbol{1}, \boldsymbol{x}), j = 1, \dotsc, J,$ with $P(L_j | \boldsymbol{R} = \boldsymbol{1}, \boldsymbol{x}) := (P(L_j = 1 | \boldsymbol{R} = \boldsymbol{1}, x_j), \dotsc, P(L_j = M(j) | \boldsymbol{R} = \boldsymbol{1}, x_j))$.

\subsection{Example} \label{sec:example}

\begin{table*}
\caption{Calculation for the pRSL Example described in Section \ref{sec:example}.}
\label{tab:example}
\centering
\begin{tabular}{cccrrrrrr}
\toprule
$\ell_1$ & $\ell_2$ & $\ell_3$ & $P(\boldsymbol{L} = \boldsymbol{\ell} | x)$ & $P(R_1=1|\boldsymbol{L} = \boldsymbol{\ell})$ & $P(R_2=1|\boldsymbol{L} = \boldsymbol{\ell})$ & $P(R_3=1|\boldsymbol{L} = \boldsymbol{\ell})$ & $P(\boldsymbol{L} = \boldsymbol{\ell} | \boldsymbol{R} = \boldsymbol{1}, x)$ \\
\midrule
$w$ &$s$ &$h$ & $0.1 \cdot 0.95 \cdot 0.5$ & $0.2$ & $0.9$ & $0$ & $0$ \\
$n$ &$s$ &$h$ & $0.4 \cdot 0.95 \cdot 0.5$ & $0.2$ & $0.1$ & $1$ & $0.0078$ \\
$o$ &$s$ &$h$ & $0.5 \cdot 0.95 \cdot 0.5$ & $0.8$ & $0.9$ & $1$ & $0.3517$ \\
$w$ &$g$ &$l$ & $0.1 \cdot 0.05 \cdot 0.2$ & $0.8$ & $0.9$ & $1$ & $0.0015$ \\
$n$ &$s$ &$c$ & $0.4 \cdot 0.95 \cdot 0.3$ & $0.8$ & $0.9$ & $1$ & $0.1688$ \\
$o$ &$s$ &$c$ & $0.5 \cdot 0.95 \cdot 0.3$ & $0.8$ & $0.9$ & $1$ & $0.2110$ \\
$\dots$ & $\dots$ & $\dots$ & $\dots$ & $\dots$ & $\dots$ & $\dots$ & $\dots$ \\
\bottomrule
\end{tabular}
\end{table*}

To better understand how pRSL combines its input beliefs with the given rules, we continue with a numerical example.

In logistics, there is a rising trend of using human activity recognition to observe workers' actions with sensors and machine learning models \citep{reining2019human}. Suppose we have three sensors, each analyzed by a machine learning model: A camera $C_1$ that can distinguish walking movement $w$ from work at a normal height $n$ and overhead work $o$. Additionally, workers wear sensor shoes $C_2$ that can distinguish standing $s$ from gait $g$, and a wristband $C_3$ that observes high $h$, centered $c$ or low $l$ hand height. All of these classifiers have imperfect accuracy, but we assume for this example's simplicity that they are well calibrated so that we can use $C$ and $L$ interchangeably. We want to connect $C_1, C_2$, and $C_3$ in order to better detect the potentially harmful overhead work. For that, we use three rules:
\[
\varphi_1: (h \wedge s) \rightarrow o~(p_1=0.8)
\]
expresses that in many cases when workers stand and have their arms up, they do overhead work. 
\[
\varphi_2: n \rightarrow (c \lor l)~(p_2=0.9)
\] 
means that during normal work the workers' hands are almost always at a centered or low height. Lastly, 
\[
\varphi_3: w \leftrightarrow g~(p_3=1)
\] 
means the camera's walking category and the sensor shoes' gait category semantically mean the same activity.

Now, consider the camera is unsure whether the worker performs overhead work or normal work, the sensor shoes are quite certain the person is standing and the wristband indicates the worker's hands are probably high: 
\begin{align*}
P(C_1|x_1) &= (w = 0.1, n = 0.4, o = 0.5), \\
P(C_2|x_2) &= (g = 0.05, s = 0.95), \\
P(C_3|x_3) &= (h = 0.5, c = 0.3, l = 0.2).
\end{align*}
Table \ref{tab:example} shows how pRSL uses the given rules to re--weight which activities seem likely and which contradict the rules and thus are less likely. For each possible $\ell$ the priors are multiplied with the probabilities of the rules given $\ell$. The normalized result is reported in the last column. From the table, we conclude that $\boldsymbol{\ell} = (o, s, h)$ is most likely. After marginalizing over each label, we obtain the updated beliefs 
\begin{align*} 
P(L_1|\boldsymbol{R}=\boldsymbol{1}, \boldsymbol{x}) &= (w = 0.01, n = 0.05, o = 0.94), \\
P(L_2|\boldsymbol{R}=\boldsymbol{1}, \boldsymbol{x}) &= (g = 0.01, s = 0.99), \\
P(L_3|\boldsymbol{R}=\boldsymbol{1}, \boldsymbol{x}) &= (h = 0.48, c = 0.31, l = 0.21).
\end{align*}
So, after combining the information of all three sensors we can be certain that the worker is performing overhead work.

This example shows how pRSL applies principles of probabilistic logic to multi--label classification and performs stacking of $C_1, C_2$, and $C_3$ while remaining interpretable. Obviously, there are more sophisticated algorithms to obtain the updated beliefs. They are portrayed hereafter.

\subsection{Multicategorical Noisy--OR} \label{sec:noisyor}

While propositional logic allows for a broad variety of statements, inference and learning are correspondingly complex. However, many formulas can be written in the form of an implication $\varphi = (A \wedge B) \rightarrow (C \lor D)$, where both the body and the head may contain multiple labels. This form can be equivalently expressed as a disjunction  $\varphi \equiv \neg A \lor \neg B \lor C \lor D$. The noisy--or gate \citep{pearl1988} is a parametrized distribution that extends the classical logic disjunction to the probabilistic setting. It is widely adapted in Bayesian networks \citep{pmlr-v115-ji20a} as it allows for faster inference and, as we will later show, efficient learning. 

In Section \ref{sec:extNor} in the appendix, we extend the noisy--or gate to our case where the output is binary, but the labels acting as inputs may be multicategorical. For a rule $R_k$, each label $m$ in each connected node $L_j$ has an inhibition probability $q_{jm}^k$ that gives the probability that the noisy--or node is not activated even though the label is active. The rule's probability previously defined in (\ref{form:crisprule}) is now softened:
\begin{align} \label{form:noisyrule}
P(R_k = 1| \boldsymbol{L} = \boldsymbol{\ell}) = 1 - \prod_{j=1}^J q^k_{j\ell_j}\,\,.
\end{align}


\subsection{Inference} \label{sec:inference}

Inference on pRSL is performed using belief propagation \citep{pearl1988}. Note that we may be interested in two kinds of queries: Either the marginal distribution of each label
\begin{align} \label{form:mpe}
P(L_j|\boldsymbol{R} = \boldsymbol{1}, \boldsymbol{x}), j = 1, \dotsc, J,
\end{align}
or the most probable explanation (MPE) 
\begin{align} \label{form:marg}
\argmax\limits_{\boldsymbol{\ell} \in \mathcal{L}} \, P(\boldsymbol{L} = \boldsymbol{\ell}|\boldsymbol{R}=\boldsymbol{1}, \boldsymbol{x}),
\end{align} 
which gives the joint setup of labels that has the highest likelihood. \citet{dembczynski2010bayes} have shown that the two queries are Bayes--optimal for different loss functions.

We implement pRSL as Bayesian network in \texttt{R} 3.6.3 using the \texttt{gRain} \citep{gRain} package that provides exact inference for marginal and MPE queries. This implementation allows processing both noisy--or and arbitrary propositional logic based rules, but is NP--complete even for bipartite noisy--or networks \citep{cooper1990computational}. To circumvent this by utilizing the noisy--or structure, we further implemented approximate algorithms for larger datasets. These loopy belief propagation \citep{10.5555/2073796.2073849} algorithms are based on the exact sum--product and max--product algorithms for acyclic networks introduced by \citet{pearl1988} and can approximate marginal and MPE queries. They achieve a runtime complexity of $O(J \cdot J_0 \cdot K)$ for marginal and $O(J \cdot J_0 \cdot K + K \cdot 2^{J_0})$ for MPE queries, where  $J_0$ is the maximum number of labels connected to a rule. An experiment on their approximation quality is found in Section~\ref{sec:approxSim}.

\subsection{Learning} \label{sec:learning}

All parts of pRSL can either be expert--given or learned from data. We assume the classifiers $C_1, \dotsc, C_J$ to be trained in advance and refer to \citet{xia2020multi} for calibrating the classifier outputs. Thus, we focus on learning the noisy--or rules determined by $\boldsymbol{q} = (q^1_{11}, \dotsc, q^K_{JM(J)})$. Our objective is to improve the predictive performance
\begin{align} \label{form:opt}
   \argmax\limits_{\boldsymbol{q}} \log(P(\boldsymbol{L} = \boldsymbol{\ell}^*|\boldsymbol{R} = \boldsymbol{1}, \boldsymbol{x})),
\end{align}
where $\boldsymbol{\ell}^*$ are the true categories of a given observation with classifier input data $\boldsymbol{x}$. 

To find the optimal rules, we apply batchwise ADAM optimization \citep{DBLP:journals/corr/KingmaB14}. In Section \ref{sec:grAllKnown} in the appendix, we build partial derivatives of (\ref{form:opt}) by $q^k_{jm}$ for all rules simultaneously. The resulting gradients can be transformed into expressions that depend only on 
\begin{align}
&P(L_j = m|R_k = 0, R_v = 1, v \neq k, \boldsymbol{x}), \label{form:gr1} \\
&P(R_k = 1|R_v = 1, v \neq k, \boldsymbol{x}), \label{form:gr2}
\end{align}
and particular values of $\boldsymbol{q}$, so that for each rule's gradient we need to perform only two marginal queries on the pRSL, thus avoiding the higher costs of MPE queries.

We can also perform gradient descent when not all true labels are known such as in zero--shot problems or incomplete datasets. The set of known labels $\{\boldsymbol{L}' = \boldsymbol{\ell}'\}$ may be different for each observation. Then, the optimization goal is
\begin{align}\label{form:partopt}
{\arg\max}_{\boldsymbol{q}} \log(P(\boldsymbol{L}' = \boldsymbol{l}'|\boldsymbol{R} = \boldsymbol{1}, \boldsymbol{x})).
\end{align}
We derive corresponding gradients in Section \ref{sec:grUnknowns} in the appendix, which is even possible for such $q^k_{jm}$ that belong to labels $L_j \not\in \boldsymbol{L}'$ with missing ground--truth. The gradients consist of (\ref{form:gr1}), (\ref{form:gr2}), specific values of $\boldsymbol{q}$, and additionally
\begin{align}
&P(L_j = m|\boldsymbol{L}' = \boldsymbol{l}', R_k = 0, R_v = 1, v \neq k, \boldsymbol{x}) \text{ and } \label{form:gr3} \\
&P(R_k = 1|\boldsymbol{L}' = \boldsymbol{l}', R_v = 1, v \neq k, \boldsymbol{x}), \label{form:gr4}
\end{align}  
i.e. two additional marginal queries are required per rule.

Overall, the number of parameters $\boldsymbol{q}$ to learn is linear in the number of rules and labels, and one step in the gradient descent can be performed in $O(K^2 \cdot J \cdot J_0)$ runtime when using approximate queries to compute the gradients. 

\subsection{Regularization} \label{sec:regularization}

The maximum number of labels connected to each rule $J_0$ plays a major rule in keeping computations bearable and rules interpretable. Thus, regularization should favor $q^k_{jm} = 1$ for all inhibition probabilities related to one label $L_j$, so that it can be disconnected from rule $R_k$. In consequence, we do not regularize the $q^k_{jm}$ directly, but their soft minimum \citep{cook2011basic} per label $L_j$ and rule $R_k$ given by
\begin{align}
s(k, j) = - \frac{1}{\alpha} \log \left( \sum\limits_{m = 1}^{M(j)} \exp(- \alpha q^k_{jm})\right),
\end{align}
where $\alpha$ controls the hardness of the soft minimum.

Regularization happens twofold: On the one hand, we apply a hard regularizer that allows only the $J_0$ label nodes $L_j$ with the lowest $s(k, j)$ to be connected to each rule $R_k$ and sets all other inhibition probabilities to $1$, thus transposing $\boldsymbol{q}$ to close but computationally efficient positions during the gradient descent. On the other hand, there is a soft regularizer to push $\boldsymbol{q}$ towards~$\boldsymbol{1}$. This soft regularizer is obtained via the duality between regularizers and Bayesian priors \citep{murphy2012machine}. Regarding $s(k, j)$ as a random variable $S$, we assume $S~\sim~\text{Beta}(\beta_1, \beta_2),$ where $0 < \beta_1, \beta_2 < 1$ are selected so that
\begin{align}
	&P(S < 0.1) = \gamma_0 \text{ and } \\
	&P(S > 0.9) = \gamma_1,
\end{align} 
with $\gamma_0 < \gamma_1$. This prior places a high mass on $1$ and a smaller mass on $0$, so that most labels are removed from the rule to lower $J_0$ and the ones used in the rule are pushed towards crisp inhibition probabilities for easier interpretability. The corresponding regularizer penalty to be added to the optimization goal (\ref{form:opt}) is
\begin{align}
\begin{split}
R(\boldsymbol{q}) =& \frac{\lambda^t}{\eta} ((\beta_1 - 1)\sum\limits_{k = 1}^K \sum\limits_{j = 1}^J \log(s(k, j) + \epsilon) + \\
& (\beta_2 - 1) \sum\limits_{k = 1}^K \sum\limits_{j = 1}^J \log(1 - s(k, j) + \epsilon)),
\end{split}
\end{align}
where $\eta$ is the normalization constant of the $\text{Beta}(\beta_1, \beta_2)$ distribution, $\epsilon$ prevents dividing by zero when differentiating and $\lambda < 1$ decreases the influence of the regularization penalty with each iteration $t$ of the gradient descent. A benefit of deriving the regularizer from a prior is that the regularizer strength is given by the normalization constant~$\eta$ and does not have to be tuned as a hyperparameter. In simulations not detailed here, $\alpha = 20$, $J_0 = 5$, $\epsilon = 10^{-4}$, and $\lambda = 0.98$ proved to be good default values.

An \texttt{R} implementation of all above methods is found online\footnote{\url{https://github.com/mkirchhof/rsl}}.

\section{Experiments} \label{sec:application}

\subsection{Approximation Quality} \label{sec:approxSim}

Though it has no quality bound, \citet{10.5555/2073796.2073849} have observed that the loopy belief propagation used for pRSL's approximate queries may work well in the bipartite noisy--or setting. However, the exact conditions are still under research \citep{jebara2013uai}. As the scalability of pRSL relies on the approximation quality, we compare exact to approximated marginal and MPE queries in various simulated pRSL models below. The code to reproduce the simulations and further analyses can be accessed online\footnote{\url{https://github.com/mkirchhof/rslSim}}.

pRSL models of three sizes were generated, each replicated $10$ times. The smallest included 5 labels and 5 rules, a medium one 10 labels and 10 rules, and the largest 30 labels and 30 rules, after which exact inference became incomputable due to its exponentially rising runtime and memory usage. The sampling procedure is detailed in Section \ref{sec:sim} in the appendix. In short, labels comprised a random amount of $2-4$ categories. Noisy--or rule related $2-5$ random categories with random inhibition probabilities.
Finally, $100$ classifier observations were simulated by drawing $\text{Dir}(1)$ distributed random variables for each label.

Marginal queries were approximated well, with correlation coefficients of $r_s = 0.995$, $r_m = 0.996$, and $r_l = 0.995$ between the exact and approximate queries averaged across all replications in the small ($r_s$), medium ($r_m$), and large ($r_l$) models. In terms of scalability, no strong reduction in quality can be seen in the rising model sizes. 

\begin{figure}
\centering
\begin{tikzpicture}[x=1pt,y=1pt]
\definecolor{fillColor}{RGB}{255,255,255}
\path[use as bounding box,fill=fillColor,fill opacity=0.00] (0,0) rectangle (216.81,144.54);
\begin{scope}
\path[clip] ( 33.60, 98.67) rectangle (216.81,131.34);
\definecolor{drawColor}{RGB}{0,0,0}

\path[draw=drawColor,line width= 1.2pt,line join=round] (155.74,101.39) -- (155.74,113.49);

\path[draw=drawColor,line width= 0.4pt,dash pattern=on 4pt off 4pt ,line join=round,line cap=round] (109.94,107.44) -- (138.78,107.44);

\path[draw=drawColor,line width= 0.4pt,dash pattern=on 4pt off 4pt ,line join=round,line cap=round] (193.06,107.44) -- (162.53,107.44);

\path[draw=drawColor,line width= 0.4pt,line join=round,line cap=round] (109.94,104.42) -- (109.94,110.47);

\path[draw=drawColor,line width= 0.4pt,line join=round,line cap=round] (193.06,104.42) -- (193.06,110.47);

\path[draw=drawColor,line width= 0.4pt,line join=round,line cap=round] (138.78,101.39) --
	(138.78,113.49) --
	(162.53,113.49) --
	(162.53,101.39) --
	(138.78,101.39);

\path[draw=drawColor,line width= 1.2pt,line join=round] (204.09,116.52) -- (204.09,128.62);

\path[draw=drawColor,line width= 0.4pt,dash pattern=on 4pt off 4pt ,line join=round,line cap=round] (191.36,122.57) -- (199.85,122.57);

\path[draw=drawColor,line width= 0.4pt,dash pattern=on 4pt off 4pt ,line join=round,line cap=round] (210.02,122.57) -- (206.63,122.57);

\path[draw=drawColor,line width= 0.4pt,line join=round,line cap=round] (191.36,119.54) -- (191.36,125.59);

\path[draw=drawColor,line width= 0.4pt,line join=round,line cap=round] (210.02,119.54) -- (210.02,125.59);

\path[draw=drawColor,line width= 0.4pt,line join=round,line cap=round] (199.85,116.52) --
	(199.85,128.62) --
	(206.63,128.62) --
	(206.63,116.52) --
	(199.85,116.52);
\end{scope}
\begin{scope}
\path[clip] (  0.00,  0.00) rectangle (216.81,144.54);
\definecolor{drawColor}{RGB}{0,0,0}

\path[draw=drawColor,line width= 0.4pt,line join=round,line cap=round] ( 33.60, 98.67) --
	(216.81, 98.67) --
	(216.81,131.34) --
	( 33.60,131.34) --
	( 33.60, 98.67);

\path[draw=drawColor,line width= 0.4pt,line join=round,line cap=round] ( 33.60,107.44) -- ( 33.60,122.57);

\path[draw=drawColor,line width= 0.4pt,line join=round,line cap=round] ( 33.60,107.44) -- ( 27.60,107.44);

\path[draw=drawColor,line width= 0.4pt,line join=round,line cap=round] ( 33.60,122.57) -- ( 27.60,122.57);

\node[text=drawColor,anchor=base east,inner sep=0pt, outer sep=0pt, scale=  1.00] at ( 21.60,104.00) {Marg};

\node[text=drawColor,anchor=base east,inner sep=0pt, outer sep=0pt, scale=  1.00] at ( 21.60,119.12) {MPE};

\path[draw=drawColor,line width= 0.4pt,line join=round,line cap=round] ( 40.39, 98.67) -- (210.02, 98.67);

\path[draw=drawColor,line width= 0.4pt,line join=round,line cap=round] ( 40.39, 98.67) -- ( 40.39, 92.67);

\path[draw=drawColor,line width= 0.4pt,line join=round,line cap=round] ( 82.80, 98.67) -- ( 82.80, 92.67);

\path[draw=drawColor,line width= 0.4pt,line join=round,line cap=round] (125.20, 98.67) -- (125.20, 92.67);

\path[draw=drawColor,line width= 0.4pt,line join=round,line cap=round] (167.61, 98.67) -- (167.61, 92.67);

\path[draw=drawColor,line width= 0.4pt,line join=round,line cap=round] (210.02, 98.67) -- (210.02, 92.67);

\node[text=drawColor,anchor=base,inner sep=0pt, outer sep=0pt, scale=  1.00] at (125.20,136.14) {Small};
\end{scope}
\begin{scope}
\path[clip] ( 33.60, 39.60) rectangle (216.81, 72.27);
\definecolor{drawColor}{RGB}{0,0,0}

\path[draw=drawColor,line width= 1.2pt,line join=round] (111.63, 42.32) -- (111.63, 54.42);

\path[draw=drawColor,line width= 0.4pt,dash pattern=on 4pt off 4pt ,line join=round,line cap=round] ( 87.88, 48.37) -- (101.46, 48.37);

\path[draw=drawColor,line width= 0.4pt,dash pattern=on 4pt off 4pt ,line join=round,line cap=round] (157.44, 48.37) -- (126.90, 48.37);

\path[draw=drawColor,line width= 0.4pt,line join=round,line cap=round] ( 87.88, 45.35) -- ( 87.88, 51.40);

\path[draw=drawColor,line width= 0.4pt,line join=round,line cap=round] (157.44, 45.35) -- (157.44, 51.40);

\path[draw=drawColor,line width= 0.4pt,line join=round,line cap=round] (101.46, 42.32) --
	(101.46, 54.42) --
	(126.90, 54.42) --
	(126.90, 42.32) --
	(101.46, 42.32);

\path[draw=drawColor,line width= 1.2pt,line join=round] (199.85, 57.45) -- (199.85, 69.55);

\path[draw=drawColor,line width= 0.4pt,dash pattern=on 4pt off 4pt ,line join=round,line cap=round] (186.27, 63.50) -- (193.06, 63.50);

\path[draw=drawColor,line width= 0.4pt,dash pattern=on 4pt off 4pt ,line join=round,line cap=round] (208.33, 63.50) -- (203.24, 63.50);

\path[draw=drawColor,line width= 0.4pt,line join=round,line cap=round] (186.27, 60.47) -- (186.27, 66.52);

\path[draw=drawColor,line width= 0.4pt,line join=round,line cap=round] (208.33, 60.47) -- (208.33, 66.52);

\path[draw=drawColor,line width= 0.4pt,line join=round,line cap=round] (193.06, 57.45) --
	(193.06, 69.55) --
	(203.24, 69.55) --
	(203.24, 57.45) --
	(193.06, 57.45);
\end{scope}
\begin{scope}
\path[clip] (  0.00,  0.00) rectangle (216.81,144.54);
\definecolor{drawColor}{RGB}{0,0,0}

\path[draw=drawColor,line width= 0.4pt,line join=round,line cap=round] ( 33.60, 39.60) --
	(216.81, 39.60) --
	(216.81, 72.27) --
	( 33.60, 72.27) --
	( 33.60, 39.60);

\path[draw=drawColor,line width= 0.4pt,line join=round,line cap=round] ( 33.60, 48.37) -- ( 33.60, 63.50);

\path[draw=drawColor,line width= 0.4pt,line join=round,line cap=round] ( 33.60, 48.37) -- ( 27.60, 48.37);

\path[draw=drawColor,line width= 0.4pt,line join=round,line cap=round] ( 33.60, 63.50) -- ( 27.60, 63.50);

\node[text=drawColor,anchor=base east,inner sep=0pt, outer sep=0pt, scale=  1.00] at ( 21.60, 44.93) {Marg};

\node[text=drawColor,anchor=base east,inner sep=0pt, outer sep=0pt, scale=  1.00] at ( 21.60, 60.05) {MPE};

\path[draw=drawColor,line width= 0.4pt,line join=round,line cap=round] ( 40.39, 39.60) -- (210.02, 39.60);

\path[draw=drawColor,line width= 0.4pt,line join=round,line cap=round] ( 40.39, 39.60) -- ( 40.39, 33.60);

\path[draw=drawColor,line width= 0.4pt,line join=round,line cap=round] ( 82.80, 39.60) -- ( 82.80, 33.60);

\path[draw=drawColor,line width= 0.4pt,line join=round,line cap=round] (125.20, 39.60) -- (125.20, 33.60);

\path[draw=drawColor,line width= 0.4pt,line join=round,line cap=round] (167.61, 39.60) -- (167.61, 33.60);

\path[draw=drawColor,line width= 0.4pt,line join=round,line cap=round] (210.02, 39.60) -- (210.02, 33.60);

\node[text=drawColor,anchor=base,inner sep=0pt, outer sep=0pt, scale=  1.00] at ( 40.39, 18.00) {0};

\node[text=drawColor,anchor=base,inner sep=0pt, outer sep=0pt, scale=  1.00] at ( 82.80, 18.00) {0.25};

\node[text=drawColor,anchor=base,inner sep=0pt, outer sep=0pt, scale=  1.00] at (125.20, 18.00) {0.5};

\node[text=drawColor,anchor=base,inner sep=0pt, outer sep=0pt, scale=  1.00] at (167.61, 18.00) {0.75};

\node[text=drawColor,anchor=base,inner sep=0pt, outer sep=0pt, scale=  1.00] at (210.02, 18.00) {1};

\node[text=drawColor,anchor=base,inner sep=0pt, outer sep=0pt, scale=  1.00] at (125.20,  3.60) {Predictions Equal to Exact MPE};

\node[text=drawColor,anchor=base,inner sep=0pt, outer sep=0pt, scale=  1.00] at (125.20, 77.07) {Medium};
\end{scope}
\end{tikzpicture}
\caption{Quality of Approximation of Exact MPEs by Approximate MPEs and Approximate Marginals.}
\label{fig:approxQuality}
\end{figure}
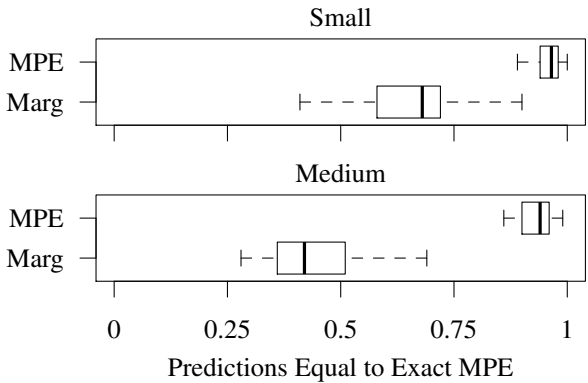

For MPE queries, the comparison could only be conducted on the small and medium datasets, as the naive implementation of exact MPE queries in the \texttt{gRain} package ruled out the large dataset. Besides approximating MPE queries by the aforementioned loopy belief propagation, they were additionally approximated in Naive--Bayes manner by combining the label--wise approximate marginal queries. Figure~\ref{fig:approxQuality} shows how often the approximate MPE queries matched their exact counterparts. It shows that approximate MPE queries clearly outperform the label--wise approximation with a median of $96.5\%$ versus $68\%$ correctly resembled MPE queries on the small and $94\%$ versus $42\%$ on the medium datasets. This underlines that questions regarding MPEs are not well answered by combining marginal queries, but need distinguished joint--label algorithms.

In summary, loopy belief propagation showed a nearly perfect approximation quality for marginal and a good approximation quality for MPE queries. The difference between the queries might be due to the higher sensitivity to small inaccuracies of the underlying max--product algorithm for MPE queries as opposed to the sum--product algorithm for marginal queries that can marginalize them out.

\subsection{Performance Benchmark}


\subsubsection{Benchmark Datasets} \label{sec:data}

In the following experiment, pRSL was compared to the state--of--the--art on six established multi--label benchmark datasets obtained from Mulan\footnote{http://mulan.sourceforge.net/datasets-mlc.html}. Each has a different domain and complexity as summarized in Table \ref{tab:data}. The three smaller datasets -- emotions, yeast, and birds -- were split into a $10$--fold cross--validation with an 8/1/1 split for train--validation--test. A $5$--fold cross--validation with a 3/1/1 split was used for the three bigger datasets. pRSL used exact queries for former and approximate queries for the latter three datasets.

\subsubsection{Comparison Methods} \label{sec:competition}

Probabilistic random forests \citep{malley2012probability} were used as binary relevance (BR) learner to transform the raw inputs into probabilistic estimates for each label individually. 
We compared pRSL to three methods that further process these initial beliefs: Two benchmark models that are based on black box decisions (NN and MLWSE) and one recent approach (BOOMER) similar in nature to pRSL.

The first method is a neural network (NN) with two hidden layers that have twice the number of labels as neurons. It uses label--wise cross--entropy as loss function and utilizes the validation data for early stopping. Second, MLWSE \citep{xia2020multi} optimizes a quadratic loss by linearly combining labels based on pairwise correlations. Third, we compare with a method already mentioned in Section \ref{sec:related}: BOOMER \citep{rapp2020learning} is a recently published soft--rule--based approach that optimizes a joint label cross--entropy loss. To maintain a fair comparison, its three hyperparameters for shrinkage, regularization strength, and number of rules were tuned via grid--search on the validation data. Last, pRSL was applied where the validation data served to find an optimal number of rules. No form of calibration was applied. Experiments could not be conducted on \citet{ding2015probabilistic}, which is similar to pRSL, due to unavailable implementation.

\subsubsection{Evaluation Measures} \label{sec:measures}

\begingroup
\setlength{\tabcolsep}{4pt} 
\begin{table}
\centering
\caption{Overview of Benchmark Datasets. Cardinality gives the Mean Number of Positive Labels per Observation. Density is the Cardinality Normalized by the Number of Labels.} \label{tab:data}
\begin{tabular}{llrrrr}
\toprule
Dataset & Domain & Labels & Cardinality & Density \\
\midrule
Emotions & Music & $6$ & $1.869$ & $0.311$\\
Yeast & Genes & $14$ & $4.237$ & $0.303$\\
Birds & Sound & $19$ & $1.014$ & $0.053$\\
Medical & Diseases & $45$ & $1.245$ & $0.028$\\
Enron & Emails & $53$ & $3.378$ & $0.064$\\
Mediamill & Newscasts & $101$ & $4.376$ & $0.043$\\
\bottomrule
\end{tabular}
\end{table}
\endgroup

\begingroup
\setlength{\tabcolsep}{4pt} 
\begin{table*}
\caption{Results of the $k$--Fold Crossvalidations. Mean $\pm$ Standard Deviation Between Folds. Best Result in Bold.}
\label{tab:benchmark}
\centering
\begin{tabular}{lcccccc}
\toprule
 & Emotions & Yeast & Birds & Medical & Enron & Mediamill \\
\midrule
\multicolumn{7}{c}{Joint Accuracy (higher = better)} \\
\midrule
BR & $0.321 \pm 0.005$ & $0.174 \pm 0.018$ & $0.510 \pm 0.004$ & $0.386 \pm 0.022$ & $0.114 \pm 0.026$ & $0.148 \pm 0.001$\\
NN & $0.309 \pm 0.075$ & $0.193 \pm 0.018$ & $0.540 \pm 0.051$ & $0.613 \pm 0.032$ & $0.120 \pm 0.020$ & $0.179 \pm 0.003$ \\
MLWSE & $0.327 \pm 0.058$ & $0.201 \pm 0.013$ & $\mathbf{0.541 \pm 0.041}$ & $0.640 \pm 0.025$ & $0.124 \pm 0.022$ & $0.151 \pm 0.001$\\
BOOMER & $0.326 \pm 0.074$ & $0.224 \pm 0.022$ & $0.527 \pm 0.024$ & $\mathbf{0.654 \pm 0.024}$ & $0.137 \pm 0.026$ & $\mathbf{0.189 \pm 0.004}$\\
pRSL & $\mathbf{0.348 \pm 0.067}$ & $\mathbf{0.236 \pm 0.015}$ & $0.507 \pm 0.032$ & $0.491 \pm 0.031$ & $\mathbf{0.153 \pm 0.020}$ & $0.149 \pm 0.002$\\
\midrule
& \multicolumn{3}{c}{Joint log--Likelihood (higher = better)} & \multicolumn{3}{c}{Label--wise log--Likelihood (higher = better)} \\
\cmidrule(r){1-4} \cmidrule(l){5-7}
BR & $-2.386 \pm 0.145$ & $-5.772 \pm 0.198$ & $-2.448 \pm 0.160$ & $-1.584 \pm 0.091$ & $-6.376 \pm 0.182$ & $-6.614 \pm 0.062$\\
NN & $-2.105 \pm 0.267$ & $-5.659 \pm 0.231$ & $-1.547 \pm 0.409$ & $\mathbf{-0.750 \pm 0.133}$ & $-6.476 \pm 0.172$ & $\mathbf{-6.016 \pm 0.038}$\\
MLWSE & $-2.140 \pm 0.202$ & $-5.512 \pm 0.227$ & $\mathbf{-1.502 \pm 0.167}$ & $-1.125 \pm 0.110$ & $\mathbf{-6.018 \pm 0.104}$ & $-6.371 \pm 0.052$\\
BOOMER & unavailable & unavailable & unavailable & unavailable & unavailable & unavailable\\
pRSL & $\mathbf{-1.839 \pm 0.273}$ & $\mathbf{-3.592 \pm 0.085}$ & $-2.458 \pm 0.156$ & $-1.565 \pm 0.120$ & $-6.479 \pm 0.242$ & $-6.532 \pm 0.061$\\
\midrule
\multicolumn{7}{c}{Label--wise Hamming Loss (lower = better)} \\
\midrule
BR & $\mathbf{0.179 \pm 0.019}$ & $0.188 \pm 0.003$ & $0.042 \pm 0.003$ & $0.017 \pm 0.001$ & $0.045 \pm 0.001$ & $0.027 \pm 0.000$\\
NN & $0.183 \pm 0.024$ & $0.191 \pm 0.005$ & $0.039 \pm 0.003$ & $0.012 \pm 0.001$ & $0.046 \pm 0.001$ & $\mathbf{0.025 \pm 0.000}$\\
MLWSE & $0.183 \pm 0.024$ & $\mathbf{0.186 \pm 0.005}$ & $\mathbf{0.037 \pm 0.002}$ & $0.011 \pm 0.001$ & $\mathbf{0.044 \pm 0.001}$ & $0.026 \pm 0.000$\\
BOOMER & $0.183 \pm 0.024$ & $0.188 \pm 0.005$ & $0.041 \pm 0.003$ & $\mathbf{0.011 \pm 0.001}$ & $0.046 \pm 0.001$ & $0.026 \pm 0.000$\\
pRSL & $0.182 \pm 0.022$ & $0.190 \pm 0.005$ & $0.043 \pm 0.002$ & $0.015 \pm 0.001$ & $0.046 \pm 0.001$ & $0.027 \pm 0.000$\\
\bottomrule
\end{tabular}
\end{table*}
\endgroup

As shown by \citet{dembczynski2012label}, measuring the percentage of misclassified observations label--wise, called Hamming loss, does not suffice in multi--label classification. Hence, following their proofs, we additionally measure the percentage of observations where all labels are correctly classified, which lays focus on the MPE estimates. To complement these crisp--decision focused metrics with a metric that judges the quality of the returned probability estimates, we measure the log--likelihood of the returned beliefs, which is the only local proper scoring rule \citep{parmigiani2009decision}. Precisely, we used the median log--likelihood of the joint labels or the individual labels in case the former was not available for a learner.

\subsubsection{Results} \label{sec:benchmark}

All code required to reproduce the experiments can be found online\footnote{\url{https://github.com/mkirchhof/rslBench}}. The results are reported in Table \ref{tab:benchmark}. As BOOMER returns no probabilistic estimates, its log--likelihood could not be computed. In general, the examined multi--label algorithms outperformed the BR baseline across all datasets and metrics, except the hamming loss on emotions. 
Interestingly, on birds and medical, BR is outperformed by a larger margin even in terms of hamming loss which is a single label metric that can be optimized without modeling multi--label dependencies \citep{dembczynski2012label}.

Each of the four multi--label algorithms showed strengths on different datasets and metrics. Except on birds, no algorithm performs best across all metrics on a fixed dataset or across all datasets on a fixed metric. In particular, MLWSE, the only linear model, performed best on birds across all metrics and three datasets in terms of hamming loss, plus a nearly indistinguishable performance with BOOMER on medical. The two soft rule based algorithms BOOMER and pRSL performed on a nearly indistinguishable level on four datasets when taking the estimation uncertainty into account, except on medical and mediamill.

pRSL showed the best performance among the benchmarked state--of--the--art methods on three out of six datasets regarding accuracy and two of the three datasets where the joint log--likelihood could be estimated. It does not show improvements on the label--wise metrics on any dataset. This may be a consequence of its internal joint--label loss function. It also increased the accuracy against the BR input beliefs on two out of the three big datasets where it relied on approximate queries. This further reinforces the claim on approximation quality made in Section \ref{sec:approxSim}. On a side note, pRSL showed no signs of overfitting (see Table \ref{tab:overfitting} in the appendix). Two hypotheses for pRSL's worse performance on birds, mediamill, and partially on medical, can be named: 

The first is these datasets' low density as seen in Table \ref{tab:data}. This issue is related to the problem of imbalanced classes, which is a common pitfall in multi--label classification, especially on these particular datasets \citep{zhang2020towards}.

\begin{figure*}
\begin{tikzpicture}[x=1pt,y=1pt]
\definecolor{fillColor}{RGB}{255,255,255}
\path[use as bounding box,fill=fillColor,fill opacity=0.00] (0,0) rectangle (102.62,115.63);
\begin{scope}
\path[clip] (  0.00,  0.00) rectangle (102.62,115.63);
\definecolor{drawColor}{RGB}{0,0,0}

\path[draw=drawColor,line width= 0.4pt,line join=round,line cap=round] ( 30.00, 45.60) --
	(102.62, 45.60) --
	(102.62,115.63) --
	( 30.00,115.63) --
	( 30.00, 45.60);
\end{scope}
\begin{scope}
\path[clip] (  0.00,  0.00) rectangle (102.62,115.63);
\definecolor{drawColor}{RGB}{0,0,0}

\path[draw=drawColor,line width= 0.4pt,line join=round,line cap=round] ( 32.69, 45.60) -- ( 99.93, 45.60);

\path[draw=drawColor,line width= 0.4pt,line join=round,line cap=round] ( 32.69, 45.60) -- ( 32.69, 39.60);

\path[draw=drawColor,line width= 0.4pt,line join=round,line cap=round] ( 39.41, 45.60) -- ( 39.41, 39.60);

\path[draw=drawColor,line width= 0.4pt,line join=round,line cap=round] ( 46.14, 45.60) -- ( 46.14, 39.60);

\path[draw=drawColor,line width= 0.4pt,line join=round,line cap=round] ( 52.86, 45.60) -- ( 52.86, 39.60);

\path[draw=drawColor,line width= 0.4pt,line join=round,line cap=round] ( 59.59, 45.60) -- ( 59.59, 39.60);

\path[draw=drawColor,line width= 0.4pt,line join=round,line cap=round] ( 66.31, 45.60) -- ( 66.31, 39.60);

\path[draw=drawColor,line width= 0.4pt,line join=round,line cap=round] ( 73.04, 45.60) -- ( 73.04, 39.60);

\path[draw=drawColor,line width= 0.4pt,line join=round,line cap=round] ( 79.76, 45.60) -- ( 79.76, 39.60);

\path[draw=drawColor,line width= 0.4pt,line join=round,line cap=round] ( 86.48, 45.60) -- ( 86.48, 39.60);

\path[draw=drawColor,line width= 0.4pt,line join=round,line cap=round] ( 93.21, 45.60) -- ( 93.21, 39.60);

\path[draw=drawColor,line width= 0.4pt,line join=round,line cap=round] ( 99.93, 45.60) -- ( 99.93, 39.60);

\node[text=drawColor,anchor=base,inner sep=0pt, outer sep=0pt, scale=  1.00] at ( 32.69, 26.07) {0};

\node[text=drawColor,anchor=base,inner sep=0pt, outer sep=0pt, scale=  1.00] at ( 52.86, 26.07) {0.3};

\node[text=drawColor,anchor=base,inner sep=0pt, outer sep=0pt, scale=  1.00] at ( 79.76, 26.07) {0.7};

\node[text=drawColor,anchor=base,inner sep=0pt, outer sep=0pt, scale=  1.00] at ( 99.93, 26.07) {1};

\path[draw=drawColor,line width= 0.4pt,line join=round,line cap=round] ( 30.00, 48.19) -- ( 30.00,113.04);

\path[draw=drawColor,line width= 0.4pt,line join=round,line cap=round] ( 30.00, 48.19) -- ( 24.00, 48.19);

\path[draw=drawColor,line width= 0.4pt,line join=round,line cap=round] ( 30.00, 54.68) -- ( 24.00, 54.68);

\path[draw=drawColor,line width= 0.4pt,line join=round,line cap=round] ( 30.00, 61.16) -- ( 24.00, 61.16);

\path[draw=drawColor,line width= 0.4pt,line join=round,line cap=round] ( 30.00, 67.65) -- ( 24.00, 67.65);

\path[draw=drawColor,line width= 0.4pt,line join=round,line cap=round] ( 30.00, 74.13) -- ( 24.00, 74.13);

\path[draw=drawColor,line width= 0.4pt,line join=round,line cap=round] ( 30.00, 80.62) -- ( 24.00, 80.62);

\path[draw=drawColor,line width= 0.4pt,line join=round,line cap=round] ( 30.00, 87.10) -- ( 24.00, 87.10);

\path[draw=drawColor,line width= 0.4pt,line join=round,line cap=round] ( 30.00, 93.58) -- ( 24.00, 93.58);

\path[draw=drawColor,line width= 0.4pt,line join=round,line cap=round] ( 30.00,100.07) -- ( 24.00,100.07);

\path[draw=drawColor,line width= 0.4pt,line join=round,line cap=round] ( 30.00,106.55) -- ( 24.00,106.55);

\path[draw=drawColor,line width= 0.4pt,line join=round,line cap=round] ( 30.00,113.04) -- ( 24.00,113.04);

\node[text=drawColor,rotate= 90.00,anchor=base,inner sep=0pt, outer sep=0pt, scale=  1.00] at ( 19.04, 48.19) {0};

\node[text=drawColor,rotate= 90.00,anchor=base,inner sep=0pt, outer sep=0pt, scale=  1.00] at ( 19.04, 67.65) {0.3};

\node[text=drawColor,rotate= 90.00,anchor=base,inner sep=0pt, outer sep=0pt, scale=  1.00] at ( 19.04, 93.58) {0.7};

\node[text=drawColor,rotate= 90.00,anchor=base,inner sep=0pt, outer sep=0pt, scale=  1.00] at ( 19.04,113.04) {1};
\end{scope}
\begin{scope}
\path[clip] ( 30.00, 45.60) rectangle (102.62,115.63);
\definecolor{drawColor}{RGB}{0,0,0}

\path[draw=drawColor,line width= 0.4pt,dash pattern=on 4pt off 4pt ,line join=round,line cap=round] ( 32.69, 48.19) --
	( 99.93,113.04);
\definecolor{drawColor}{RGB}{141,211,199}

\path[draw=drawColor,line width= 0.4pt,line join=round,line cap=round] ( 32.69, 48.79) --
	( 39.41, 50.30) --
	( 46.14, 57.70) --
	( 52.86, 62.86) --
	( 59.59, 77.14) --
	( 66.31, 86.02) --
	( 73.04,101.25) --
	( 79.76, 98.48) --
	( 86.48,105.18) --
	( 93.21,108.99) --
	( 99.93,113.04);
\definecolor{drawColor}{RGB}{255,255,179}

\path[draw=drawColor,line width= 0.4pt,line join=round,line cap=round] ( 32.69, 48.86) --
	( 39.41, 53.11) --
	( 46.14, 54.23) --
	( 52.86, 72.60) --
	( 59.59, 74.57) --
	( 66.31, 85.60) --
	( 73.04, 90.34) --
	( 79.76, 98.63) --
	( 86.48,101.92) --
	( 93.21,108.99) --
	( 99.93,113.04);
\definecolor{drawColor}{RGB}{190,186,218}

\path[draw=drawColor,line width= 0.4pt,line join=round,line cap=round] ( 32.69, 48.87) --
	( 39.41, 53.22) --
	( 46.14, 57.08) --
	( 52.86, 63.59) --
	( 59.59, 70.16) --
	( 66.31, 77.48) --
	( 73.04, 94.69) --
	( 79.76, 95.24) --
	( 86.48,102.23) --
	( 93.21,107.63) --
	( 99.93,113.04);
\definecolor{drawColor}{RGB}{251,128,114}

\path[draw=drawColor,line width= 0.4pt,line join=round,line cap=round] ( 32.69, 48.84) --
	( 39.41, 51.83) --
	( 46.14, 55.12) --
	( 52.86, 69.35) --
	( 59.59, 67.33) --
	( 66.31, 83.46) --
	( 73.04, 95.03) --
	( 79.76, 95.67) --
	( 86.48,106.76) --
	( 93.21,103.77) --
	( 99.93,113.04);
\definecolor{drawColor}{RGB}{128,177,211}

\path[draw=drawColor,line width= 0.4pt,line join=round,line cap=round] ( 32.69, 48.19) --
	( 39.41, 51.65) --
	( 46.14, 58.89) --
	( 52.86, 65.45) --
	( 59.59, 72.08) --
	( 66.31, 88.10) --
	( 73.04, 90.75) --
	( 79.76,104.20) --
	( 86.48,103.21) --
	( 93.21,113.04);
\definecolor{drawColor}{RGB}{253,180,98}

\path[draw=drawColor,line width= 0.4pt,line join=round,line cap=round] ( 32.69, 48.19) --
	( 39.41, 51.12) --
	( 46.14, 56.05) --
	( 52.86, 70.07) --
	( 59.59, 75.57) --
	( 66.31, 92.01) --
	( 73.04, 95.97) --
	( 79.76, 98.48) --
	( 86.48, 99.39) --
	( 93.21,108.41) --
	( 99.93,113.04);
\definecolor{drawColor}{RGB}{179,222,105}

\path[draw=drawColor,line width= 0.4pt,line join=round,line cap=round] ( 32.69, 49.65) --
	( 39.41, 51.01) --
	( 46.14, 57.74) --
	( 52.86, 62.90) --
	( 59.59, 70.89) --
	( 66.31, 89.79) --
	( 73.04, 97.09) --
	( 79.76, 99.05) --
	( 86.48,100.69) --
	( 93.21,113.04);
\definecolor{drawColor}{RGB}{252,205,229}

\path[draw=drawColor,line width= 0.4pt,line join=round,line cap=round] ( 32.69, 48.84) --
	( 39.41, 51.95) --
	( 46.14, 57.46) --
	( 52.86, 61.61) --
	( 59.59, 70.17) --
	( 66.31, 78.71) --
	( 73.04, 93.48) --
	( 79.76,103.25) --
	( 86.48,101.92) --
	( 93.21,113.04) --
	( 99.93,113.04);
\definecolor{drawColor}{gray}{0.85}

\path[draw=drawColor,line width= 0.4pt,line join=round,line cap=round] ( 32.69, 49.44) --
	( 39.41, 54.92) --
	( 46.14, 64.14) --
	( 52.86, 61.80) --
	( 59.59, 73.41) --
	( 66.31, 80.62) --
	( 73.04, 93.30) --
	( 79.76, 93.09) --
	( 86.48,102.23) --
	( 93.21,113.04) --
	( 99.93,113.04);
\definecolor{drawColor}{RGB}{188,128,189}

\path[draw=drawColor,line width= 0.4pt,line join=round,line cap=round] ( 32.69, 48.86) --
	( 39.41, 52.55) --
	( 46.14, 57.78) --
	( 52.86, 60.88) --
	( 59.59, 73.67) --
	( 66.31, 86.25) --
	( 73.04, 89.26) --
	( 79.76, 97.19) --
	( 86.48,106.55) --
	( 93.21,113.04) --
	( 99.93,113.04);
\end{scope}
\begin{scope}
\path[clip] (  0.00,  0.00) rectangle (102.62,115.63);
\definecolor{drawColor}{RGB}{0,0,0}

\node[text=drawColor,anchor=base,inner sep=0pt, outer sep=0pt, scale=  1.00] at ( 66.31, 13.20) {Prediction};

\node[text=drawColor,rotate= 90.00,anchor=base,inner sep=0pt, outer sep=0pt, scale=  1.00] at (  8.40, 80.62) {Positive Labels};

\node[text=drawColor,anchor=base,inner sep=0pt, outer sep=0pt, scale=  1.00] at ( 66.31,  2.40) {Emotions};
\end{scope}
\end{tikzpicture}\hspace{-0.2mm}
\begin{tikzpicture}[x=1pt,y=1pt]
\definecolor{fillColor}{RGB}{255,255,255}
\path[use as bounding box,fill=fillColor,fill opacity=0.00] (0,0) rectangle ( 72.27,115.63);
\begin{scope}
\path[clip] (  0.00,  0.00) rectangle ( 72.27,115.63);
\definecolor{drawColor}{RGB}{0,0,0}

\path[draw=drawColor,line width= 0.4pt,line join=round,line cap=round] (  0.00, 45.60) --
	( 72.27, 45.60) --
	( 72.27,115.63) --
	(  0.00,115.63) --
	(  0.00, 45.60);
\end{scope}
\begin{scope}
\path[clip] (  0.00,  0.00) rectangle ( 72.27,115.63);
\definecolor{drawColor}{RGB}{0,0,0}

\path[draw=drawColor,line width= 0.4pt,line join=round,line cap=round] (  2.68, 45.60) -- ( 69.59, 45.60);

\path[draw=drawColor,line width= 0.4pt,line join=round,line cap=round] (  2.68, 45.60) -- (  2.68, 39.60);

\path[draw=drawColor,line width= 0.4pt,line join=round,line cap=round] (  9.37, 45.60) -- (  9.37, 39.60);

\path[draw=drawColor,line width= 0.4pt,line join=round,line cap=round] ( 16.06, 45.60) -- ( 16.06, 39.60);

\path[draw=drawColor,line width= 0.4pt,line join=round,line cap=round] ( 22.75, 45.60) -- ( 22.75, 39.60);

\path[draw=drawColor,line width= 0.4pt,line join=round,line cap=round] ( 29.44, 45.60) -- ( 29.44, 39.60);

\path[draw=drawColor,line width= 0.4pt,line join=round,line cap=round] ( 36.13, 45.60) -- ( 36.13, 39.60);

\path[draw=drawColor,line width= 0.4pt,line join=round,line cap=round] ( 42.83, 45.60) -- ( 42.83, 39.60);

\path[draw=drawColor,line width= 0.4pt,line join=round,line cap=round] ( 49.52, 45.60) -- ( 49.52, 39.60);

\path[draw=drawColor,line width= 0.4pt,line join=round,line cap=round] ( 56.21, 45.60) -- ( 56.21, 39.60);

\path[draw=drawColor,line width= 0.4pt,line join=round,line cap=round] ( 62.90, 45.60) -- ( 62.90, 39.60);

\path[draw=drawColor,line width= 0.4pt,line join=round,line cap=round] ( 69.59, 45.60) -- ( 69.59, 39.60);

\node[text=drawColor,anchor=base,inner sep=0pt, outer sep=0pt, scale=  1.00] at (  2.68, 26.07) {0};

\node[text=drawColor,anchor=base,inner sep=0pt, outer sep=0pt, scale=  1.00] at ( 22.75, 26.07) {0.3};

\node[text=drawColor,anchor=base,inner sep=0pt, outer sep=0pt, scale=  1.00] at ( 49.52, 26.07) {0.7};

\node[text=drawColor,anchor=base,inner sep=0pt, outer sep=0pt, scale=  1.00] at ( 69.59, 26.07) {1};
\end{scope}
\begin{scope}
\path[clip] (  0.00, 45.60) rectangle ( 72.27,115.63);
\definecolor{drawColor}{RGB}{0,0,0}

\path[draw=drawColor,line width= 0.4pt,dash pattern=on 4pt off 4pt ,line join=round,line cap=round] (  2.68, 48.19) --
	( 69.59,113.04);
\definecolor{drawColor}{RGB}{141,211,199}

\path[draw=drawColor,line width= 0.4pt,line join=round,line cap=round] (  2.68, 49.22) --
	(  9.37, 52.98) --
	( 16.06, 57.46) --
	( 22.75, 64.42) --
	( 29.44, 73.37) --
	( 36.13, 85.18) --
	( 42.83, 91.96) --
	( 49.52, 94.99) --
	( 56.21,100.37) --
	( 62.90,110.39) --
	( 69.59,113.04);
\definecolor{drawColor}{RGB}{255,255,179}

\path[draw=drawColor,line width= 0.4pt,line join=round,line cap=round] (  2.68, 49.93) --
	(  9.37, 53.93) --
	( 16.06, 57.85) --
	( 22.75, 64.82) --
	( 29.44, 74.98) --
	( 36.13, 82.00) --
	( 42.83, 95.98) --
	( 49.52, 96.20) --
	( 56.21,103.12) --
	( 62.90,111.67) --
	( 69.59,113.04);
\definecolor{drawColor}{RGB}{190,186,218}

\path[draw=drawColor,line width= 0.4pt,line join=round,line cap=round] (  2.68, 49.13) --
	(  9.37, 53.03) --
	( 16.06, 58.41) --
	( 22.75, 65.17) --
	( 29.44, 73.08) --
	( 36.13, 85.90) --
	( 42.83, 94.01) --
	( 49.52, 95.65) --
	( 56.21,101.04) --
	( 62.90,110.09) --
	( 69.59,113.04);
\definecolor{drawColor}{RGB}{251,128,114}

\path[draw=drawColor,line width= 0.4pt,line join=round,line cap=round] (  2.68, 49.47) --
	(  9.37, 53.42) --
	( 16.06, 59.31) --
	( 22.75, 64.20) --
	( 29.44, 70.28) --
	( 36.13, 84.44) --
	( 42.83, 94.68) --
	( 49.52, 93.11) --
	( 56.21,102.85) --
	( 62.90,113.04) --
	( 69.59,113.04);
\definecolor{drawColor}{RGB}{128,177,211}

\path[draw=drawColor,line width= 0.4pt,line join=round,line cap=round] (  2.68, 49.01) --
	(  9.37, 52.88) --
	( 16.06, 57.78) --
	( 22.75, 64.46) --
	( 29.44, 75.09) --
	( 36.13, 85.01) --
	( 42.83, 93.69) --
	( 49.52, 96.35) --
	( 56.21,104.74) --
	( 62.90,108.85) --
	( 69.59,109.95);
\definecolor{drawColor}{RGB}{253,180,98}

\path[draw=drawColor,line width= 0.4pt,line join=round,line cap=round] (  2.68, 49.70) --
	(  9.37, 53.33) --
	( 16.06, 57.25) --
	( 22.75, 63.57) --
	( 29.44, 74.39) --
	( 36.13, 90.35) --
	( 42.83, 93.17) --
	( 49.52, 92.03) --
	( 56.21,100.77) --
	( 62.90,108.68) --
	( 69.59,113.04);
\definecolor{drawColor}{RGB}{179,222,105}

\path[draw=drawColor,line width= 0.4pt,line join=round,line cap=round] (  2.68, 49.07) --
	(  9.37, 53.19) --
	( 16.06, 57.82) --
	( 22.75, 64.42) --
	( 29.44, 75.93) --
	( 36.13, 82.42) --
	( 42.83, 92.84) --
	( 49.52, 94.76) --
	( 56.21,104.51) --
	( 62.90,112.28) --
	( 69.59,113.04);
\definecolor{drawColor}{RGB}{252,205,229}

\path[draw=drawColor,line width= 0.4pt,line join=round,line cap=round] (  2.68, 48.88) --
	(  9.37, 53.01) --
	( 16.06, 57.96) --
	( 22.75, 66.47) --
	( 29.44, 72.03) --
	( 36.13, 82.46) --
	( 42.83, 90.33) --
	( 49.52, 94.89) --
	( 56.21,102.60) --
	( 62.90,111.09) --
	( 69.59,113.04);
\definecolor{drawColor}{gray}{0.85}

\path[draw=drawColor,line width= 0.4pt,line join=round,line cap=round] (  2.68, 50.43) --
	(  9.37, 52.86) --
	( 16.06, 56.44) --
	( 22.75, 65.74) --
	( 29.44, 71.70) --
	( 36.13, 85.37) --
	( 42.83, 91.19) --
	( 49.52, 94.26) --
	( 56.21,103.45) --
	( 62.90,108.75) --
	( 69.59,113.04);
\definecolor{drawColor}{RGB}{188,128,189}

\path[draw=drawColor,line width= 0.4pt,line join=round,line cap=round] (  2.68, 49.43) --
	(  9.37, 53.82) --
	( 16.06, 57.92) --
	( 22.75, 64.30) --
	( 29.44, 74.24) --
	( 36.13, 83.97) --
	( 42.83, 92.71) --
	( 49.52, 96.67) --
	( 56.21,102.85) --
	( 62.90,109.56) --
	( 69.59,110.34);
\end{scope}
\begin{scope}
\path[clip] (  0.00,  0.00) rectangle ( 72.27,115.63);
\definecolor{drawColor}{RGB}{0,0,0}

\node[text=drawColor,anchor=base,inner sep=0pt, outer sep=0pt, scale=  1.00] at ( 36.13, 13.20) {Prediction};

\node[text=drawColor,anchor=base,inner sep=0pt, outer sep=0pt, scale=  1.00] at ( 36.13,  2.40) {Yeast};
\end{scope}
\end{tikzpicture}\hspace{-0.2mm}
\begin{tikzpicture}[x=1pt,y=1pt]
\definecolor{fillColor}{RGB}{255,255,255}
\path[use as bounding box,fill=fillColor,fill opacity=0.00] (0,0) rectangle ( 72.27,115.63);
\begin{scope}
\path[clip] (  0.00,  0.00) rectangle ( 72.27,115.63);
\definecolor{drawColor}{RGB}{0,0,0}

\path[draw=drawColor,line width= 0.4pt,line join=round,line cap=round] (  0.00, 45.60) --
	( 72.27, 45.60) --
	( 72.27,115.63) --
	(  0.00,115.63) --
	(  0.00, 45.60);
\end{scope}
\begin{scope}
\path[clip] (  0.00,  0.00) rectangle ( 72.27,115.63);
\definecolor{drawColor}{RGB}{0,0,0}

\path[draw=drawColor,line width= 0.4pt,line join=round,line cap=round] (  2.68, 45.60) -- ( 69.59, 45.60);

\path[draw=drawColor,line width= 0.4pt,line join=round,line cap=round] (  2.68, 45.60) -- (  2.68, 39.60);

\path[draw=drawColor,line width= 0.4pt,line join=round,line cap=round] (  9.37, 45.60) -- (  9.37, 39.60);

\path[draw=drawColor,line width= 0.4pt,line join=round,line cap=round] ( 16.06, 45.60) -- ( 16.06, 39.60);

\path[draw=drawColor,line width= 0.4pt,line join=round,line cap=round] ( 22.75, 45.60) -- ( 22.75, 39.60);

\path[draw=drawColor,line width= 0.4pt,line join=round,line cap=round] ( 29.44, 45.60) -- ( 29.44, 39.60);

\path[draw=drawColor,line width= 0.4pt,line join=round,line cap=round] ( 36.13, 45.60) -- ( 36.13, 39.60);

\path[draw=drawColor,line width= 0.4pt,line join=round,line cap=round] ( 42.83, 45.60) -- ( 42.83, 39.60);

\path[draw=drawColor,line width= 0.4pt,line join=round,line cap=round] ( 49.52, 45.60) -- ( 49.52, 39.60);

\path[draw=drawColor,line width= 0.4pt,line join=round,line cap=round] ( 56.21, 45.60) -- ( 56.21, 39.60);

\path[draw=drawColor,line width= 0.4pt,line join=round,line cap=round] ( 62.90, 45.60) -- ( 62.90, 39.60);

\path[draw=drawColor,line width= 0.4pt,line join=round,line cap=round] ( 69.59, 45.60) -- ( 69.59, 39.60);

\node[text=drawColor,anchor=base,inner sep=0pt, outer sep=0pt, scale=  1.00] at (  2.68, 26.07) {0};

\node[text=drawColor,anchor=base,inner sep=0pt, outer sep=0pt, scale=  1.00] at ( 22.75, 26.07) {0.3};

\node[text=drawColor,anchor=base,inner sep=0pt, outer sep=0pt, scale=  1.00] at ( 49.52, 26.07) {0.7};

\node[text=drawColor,anchor=base,inner sep=0pt, outer sep=0pt, scale=  1.00] at ( 69.59, 26.07) {1};
\end{scope}
\begin{scope}
\path[clip] (  0.00, 45.60) rectangle ( 72.27,115.63);
\definecolor{drawColor}{RGB}{0,0,0}

\path[draw=drawColor,line width= 0.4pt,dash pattern=on 4pt off 4pt ,line join=round,line cap=round] (  2.68, 48.19) --
	( 69.59,113.04);
\definecolor{drawColor}{RGB}{141,211,199}

\path[draw=drawColor,line width= 0.4pt,line join=round,line cap=round] (  2.68, 48.97) --
	(  9.37, 51.66) --
	( 16.06, 59.17) --
	( 22.75, 75.21) --
	( 29.44, 90.34) --
	( 36.13, 95.35) --
	( 42.83,100.88) --
	( 49.52,113.04) --
	( 56.21,113.04);
\definecolor{drawColor}{RGB}{255,255,179}

\path[draw=drawColor,line width= 0.4pt,line join=round,line cap=round] (  2.68, 48.84) --
	(  9.37, 51.48) --
	( 16.06, 58.33) --
	( 22.75, 68.24) --
	( 29.44, 74.13) --
	( 36.13, 93.09) --
	( 42.83,113.04) --
	( 49.52,103.77) --
	( 56.21,113.04) --
	( 62.90,113.04);
\definecolor{drawColor}{RGB}{190,186,218}

\path[draw=drawColor,line width= 0.4pt,line join=round,line cap=round] (  2.68, 48.86) --
	(  9.37, 52.64) --
	( 16.06, 59.68) --
	( 22.75, 66.90) --
	( 29.44,101.76) --
	( 36.13, 99.14) --
	( 42.83,104.93) --
	( 49.52,100.07) --
	( 56.21,102.23) --
	( 62.90, 96.83);
\definecolor{drawColor}{RGB}{251,128,114}

\path[draw=drawColor,line width= 0.4pt,line join=round,line cap=round] (  2.68, 48.57) --
	(  9.37, 51.65) --
	( 16.06, 57.84) --
	( 22.75, 66.40) --
	( 29.44, 97.39) --
	( 36.13,100.69) --
	( 42.83,104.39) --
	( 49.52,103.77) --
	( 56.21,113.04) --
	( 62.90,113.04);
\definecolor{drawColor}{RGB}{128,177,211}

\path[draw=drawColor,line width= 0.4pt,line join=round,line cap=round] (  2.68, 48.62) --
	(  9.37, 52.02) --
	( 16.06, 59.73) --
	( 22.75, 73.22) --
	( 29.44, 79.42) --
	( 36.13, 93.58) --
	( 42.83,106.55) --
	( 49.52, 96.83) --
	( 56.21,113.04) --
	( 62.90,113.04);
\definecolor{drawColor}{RGB}{253,180,98}

\path[draw=drawColor,line width= 0.4pt,line join=round,line cap=round] (  2.68, 48.63) --
	(  9.37, 51.30) --
	( 16.06, 59.76) --
	( 22.75, 69.34) --
	( 29.44, 83.75) --
	( 36.13,102.80) --
	( 42.83,102.23) --
	( 49.52,102.23) --
	( 56.21,113.04) --
	( 62.90,113.04);
\definecolor{drawColor}{RGB}{179,222,105}

\path[draw=drawColor,line width= 0.4pt,line join=round,line cap=round] (  2.68, 48.53) --
	(  9.37, 51.57) --
	( 16.06, 60.46) --
	( 22.75, 68.37) --
	( 29.44, 95.97) --
	( 36.13,105.83) --
	( 42.83,107.63) --
	( 49.52,113.04) --
	( 56.21,113.04);
\definecolor{drawColor}{RGB}{252,205,229}

\path[draw=drawColor,line width= 0.4pt,line join=round,line cap=round] (  2.68, 48.78) --
	(  9.37, 52.02) --
	( 16.06, 60.93) --
	( 22.75, 77.91) --
	( 29.44,102.23) --
	( 36.13,100.07) --
	( 42.83,113.04) --
	( 49.52,105.83) --
	( 56.21, 80.62) --
	( 62.90,113.04);
\definecolor{drawColor}{gray}{0.85}

\path[draw=drawColor,line width= 0.4pt,line join=round,line cap=round] (  2.68, 48.65) --
	(  9.37, 50.92) --
	( 16.06, 58.55) --
	( 22.75, 67.06) --
	( 29.44, 88.34) --
	( 36.13, 98.63) --
	( 42.83,103.77) --
	( 49.52, 98.07) --
	( 56.21,113.04) --
	( 62.90,113.04);
\definecolor{drawColor}{RGB}{188,128,189}

\path[draw=drawColor,line width= 0.4pt,line join=round,line cap=round] (  2.68, 48.94) --
	(  9.37, 51.20) --
	( 16.06, 53.77) --
	( 22.75, 76.39) --
	( 29.44, 85.60) --
	( 36.13, 91.42) --
	( 42.83,105.41) --
	( 49.52,102.23) --
	( 56.21,113.04) --
	( 62.90,113.04);
\end{scope}
\begin{scope}
\path[clip] (  0.00,  0.00) rectangle ( 72.27,115.63);
\definecolor{drawColor}{RGB}{0,0,0}

\node[text=drawColor,anchor=base,inner sep=0pt, outer sep=0pt, scale=  1.00] at ( 36.13, 13.20) {Prediction};

\node[text=drawColor,anchor=base,inner sep=0pt, outer sep=0pt, scale=  1.00] at ( 36.13,  2.40) {Birds};
\end{scope}
\end{tikzpicture}\hspace{-0.2mm}
\begin{tikzpicture}[x=1pt,y=1pt]
\definecolor{fillColor}{RGB}{255,255,255}
\path[use as bounding box,fill=fillColor,fill opacity=0.00] (0,0) rectangle ( 72.27,115.63);
\begin{scope}
\path[clip] (  0.00,  0.00) rectangle ( 72.27,115.63);
\definecolor{drawColor}{RGB}{0,0,0}

\path[draw=drawColor,line width= 0.4pt,line join=round,line cap=round] (  0.00, 45.60) --
	( 72.27, 45.60) --
	( 72.27,115.63) --
	(  0.00,115.63) --
	(  0.00, 45.60);
\end{scope}
\begin{scope}
\path[clip] (  0.00,  0.00) rectangle ( 72.27,115.63);
\definecolor{drawColor}{RGB}{0,0,0}

\path[draw=drawColor,line width= 0.4pt,line join=round,line cap=round] (  2.68, 45.60) -- ( 69.59, 45.60);

\path[draw=drawColor,line width= 0.4pt,line join=round,line cap=round] (  2.68, 45.60) -- (  2.68, 39.60);

\path[draw=drawColor,line width= 0.4pt,line join=round,line cap=round] (  9.37, 45.60) -- (  9.37, 39.60);

\path[draw=drawColor,line width= 0.4pt,line join=round,line cap=round] ( 16.06, 45.60) -- ( 16.06, 39.60);

\path[draw=drawColor,line width= 0.4pt,line join=round,line cap=round] ( 22.75, 45.60) -- ( 22.75, 39.60);

\path[draw=drawColor,line width= 0.4pt,line join=round,line cap=round] ( 29.44, 45.60) -- ( 29.44, 39.60);

\path[draw=drawColor,line width= 0.4pt,line join=round,line cap=round] ( 36.13, 45.60) -- ( 36.13, 39.60);

\path[draw=drawColor,line width= 0.4pt,line join=round,line cap=round] ( 42.83, 45.60) -- ( 42.83, 39.60);

\path[draw=drawColor,line width= 0.4pt,line join=round,line cap=round] ( 49.52, 45.60) -- ( 49.52, 39.60);

\path[draw=drawColor,line width= 0.4pt,line join=round,line cap=round] ( 56.21, 45.60) -- ( 56.21, 39.60);

\path[draw=drawColor,line width= 0.4pt,line join=round,line cap=round] ( 62.90, 45.60) -- ( 62.90, 39.60);

\path[draw=drawColor,line width= 0.4pt,line join=round,line cap=round] ( 69.59, 45.60) -- ( 69.59, 39.60);

\node[text=drawColor,anchor=base,inner sep=0pt, outer sep=0pt, scale=  1.00] at (  2.68, 26.07) {0};

\node[text=drawColor,anchor=base,inner sep=0pt, outer sep=0pt, scale=  1.00] at ( 22.75, 26.07) {0.3};

\node[text=drawColor,anchor=base,inner sep=0pt, outer sep=0pt, scale=  1.00] at ( 49.52, 26.07) {0.7};

\node[text=drawColor,anchor=base,inner sep=0pt, outer sep=0pt, scale=  1.00] at ( 69.59, 26.07) {1};
\end{scope}
\begin{scope}
\path[clip] (  0.00, 45.60) rectangle ( 72.27,115.63);
\definecolor{drawColor}{RGB}{0,0,0}

\path[draw=drawColor,line width= 0.4pt,dash pattern=on 4pt off 4pt ,line join=round,line cap=round] (  2.68, 48.19) --
	( 69.59,113.04);
\definecolor{drawColor}{RGB}{141,211,199}

\path[draw=drawColor,line width= 0.4pt,line join=round,line cap=round] (  2.68, 48.28) --
	(  9.37, 50.67) --
	( 16.06, 59.52) --
	( 22.75, 75.63) --
	( 29.44, 86.70) --
	( 36.13,103.20) --
	( 42.83,104.82) --
	( 49.52,108.34) --
	( 56.21,110.09) --
	( 62.90,113.04);
\definecolor{drawColor}{RGB}{255,255,179}

\path[draw=drawColor,line width= 0.4pt,line join=round,line cap=round] (  2.68, 48.29) --
	(  9.37, 50.63) --
	( 16.06, 60.93) --
	( 22.75, 72.85) --
	( 29.44, 96.41) --
	( 36.13,100.47) --
	( 42.83,104.28) --
	( 49.52,107.63) --
	( 56.21,106.28) --
	( 62.90,113.04);
\definecolor{drawColor}{RGB}{190,186,218}

\path[draw=drawColor,line width= 0.4pt,line join=round,line cap=round] (  2.68, 48.31) --
	(  9.37, 50.73) --
	( 16.06, 56.88) --
	( 22.75, 71.92) --
	( 29.44, 93.21) --
	( 36.13, 98.77) --
	( 42.83, 98.53) --
	( 49.52,106.90) --
	( 56.21,107.14);
\definecolor{drawColor}{RGB}{251,128,114}

\path[draw=drawColor,line width= 0.4pt,line join=round,line cap=round] (  2.68, 48.28) --
	(  9.37, 49.88) --
	( 16.06, 57.21) --
	( 22.75, 73.94) --
	( 29.44, 92.41) --
	( 36.13, 99.62) --
	( 42.83,106.36) --
	( 49.52,107.14) --
	( 56.21,111.53) --
	( 62.90,113.04);
\definecolor{drawColor}{RGB}{128,177,211}

\path[draw=drawColor,line width= 0.4pt,line join=round,line cap=round] (  2.68, 48.29) --
	(  9.37, 51.36) --
	( 16.06, 56.23) --
	( 22.75, 69.81) --
	( 29.44, 90.48) --
	( 36.13,102.60) --
	( 42.83,103.97) --
	( 49.52,107.85) --
	( 56.21,108.41) --
	( 62.90,113.04);
\end{scope}
\begin{scope}
\path[clip] (  0.00,  0.00) rectangle ( 72.27,115.63);
\definecolor{drawColor}{RGB}{0,0,0}

\node[text=drawColor,anchor=base,inner sep=0pt, outer sep=0pt, scale=  1.00] at ( 36.13, 13.20) {Prediction};

\node[text=drawColor,anchor=base,inner sep=0pt, outer sep=0pt, scale=  1.00] at ( 36.13,  2.40) {Medical};
\end{scope}
\end{tikzpicture}\hspace{-0.2mm}
\begin{tikzpicture}[x=1pt,y=1pt]
\definecolor{fillColor}{RGB}{255,255,255}
\path[use as bounding box,fill=fillColor,fill opacity=0.00] (0,0) rectangle ( 72.27,115.63);
\begin{scope}
\path[clip] (  0.00,  0.00) rectangle ( 72.27,115.63);
\definecolor{drawColor}{RGB}{0,0,0}

\path[draw=drawColor,line width= 0.4pt,line join=round,line cap=round] (  0.00, 45.60) --
	( 72.27, 45.60) --
	( 72.27,115.63) --
	(  0.00,115.63) --
	(  0.00, 45.60);
\end{scope}
\begin{scope}
\path[clip] (  0.00,  0.00) rectangle ( 72.27,115.63);
\definecolor{drawColor}{RGB}{0,0,0}

\path[draw=drawColor,line width= 0.4pt,line join=round,line cap=round] (  2.68, 45.60) -- ( 69.59, 45.60);

\path[draw=drawColor,line width= 0.4pt,line join=round,line cap=round] (  2.68, 45.60) -- (  2.68, 39.60);

\path[draw=drawColor,line width= 0.4pt,line join=round,line cap=round] (  9.37, 45.60) -- (  9.37, 39.60);

\path[draw=drawColor,line width= 0.4pt,line join=round,line cap=round] ( 16.06, 45.60) -- ( 16.06, 39.60);

\path[draw=drawColor,line width= 0.4pt,line join=round,line cap=round] ( 22.75, 45.60) -- ( 22.75, 39.60);

\path[draw=drawColor,line width= 0.4pt,line join=round,line cap=round] ( 29.44, 45.60) -- ( 29.44, 39.60);

\path[draw=drawColor,line width= 0.4pt,line join=round,line cap=round] ( 36.13, 45.60) -- ( 36.13, 39.60);

\path[draw=drawColor,line width= 0.4pt,line join=round,line cap=round] ( 42.83, 45.60) -- ( 42.83, 39.60);

\path[draw=drawColor,line width= 0.4pt,line join=round,line cap=round] ( 49.52, 45.60) -- ( 49.52, 39.60);

\path[draw=drawColor,line width= 0.4pt,line join=round,line cap=round] ( 56.21, 45.60) -- ( 56.21, 39.60);

\path[draw=drawColor,line width= 0.4pt,line join=round,line cap=round] ( 62.90, 45.60) -- ( 62.90, 39.60);

\path[draw=drawColor,line width= 0.4pt,line join=round,line cap=round] ( 69.59, 45.60) -- ( 69.59, 39.60);

\node[text=drawColor,anchor=base,inner sep=0pt, outer sep=0pt, scale=  1.00] at (  2.68, 26.07) {0};

\node[text=drawColor,anchor=base,inner sep=0pt, outer sep=0pt, scale=  1.00] at ( 22.75, 26.07) {0.3};

\node[text=drawColor,anchor=base,inner sep=0pt, outer sep=0pt, scale=  1.00] at ( 49.52, 26.07) {0.7};

\node[text=drawColor,anchor=base,inner sep=0pt, outer sep=0pt, scale=  1.00] at ( 69.59, 26.07) {1};
\end{scope}
\begin{scope}
\path[clip] (  0.00, 45.60) rectangle ( 72.27,115.63);
\definecolor{drawColor}{RGB}{0,0,0}

\path[draw=drawColor,line width= 0.4pt,dash pattern=on 4pt off 4pt ,line join=round,line cap=round] (  2.68, 48.19) --
	( 69.59,113.04);
\definecolor{drawColor}{RGB}{141,211,199}

\path[draw=drawColor,line width= 0.4pt,line join=round,line cap=round] (  2.68, 48.86) --
	(  9.37, 52.75) --
	( 16.06, 58.03) --
	( 22.75, 67.55) --
	( 29.44, 74.27) --
	( 36.13, 82.23) --
	( 42.83, 91.64) --
	( 49.52, 95.68) --
	( 56.21,102.99) --
	( 62.90,107.59) --
	( 69.59,108.23);
\definecolor{drawColor}{RGB}{255,255,179}

\path[draw=drawColor,line width= 0.4pt,line join=round,line cap=round] (  2.68, 48.95) --
	(  9.37, 51.95) --
	( 16.06, 56.91) --
	( 22.75, 64.82) --
	( 29.44, 73.37) --
	( 36.13, 86.50) --
	( 42.83, 89.10) --
	( 49.52, 95.34) --
	( 56.21,101.45) --
	( 62.90,107.21) --
	( 69.59,113.04);
\definecolor{drawColor}{RGB}{190,186,218}

\path[draw=drawColor,line width= 0.4pt,line join=round,line cap=round] (  2.68, 48.89) --
	(  9.37, 52.05) --
	( 16.06, 58.39) --
	( 22.75, 65.89) --
	( 29.44, 75.43) --
	( 36.13, 88.40) --
	( 42.83, 90.95) --
	( 49.52, 98.14) --
	( 56.21, 96.55) --
	( 62.90,105.56) --
	( 69.59,110.64);
\definecolor{drawColor}{RGB}{251,128,114}

\path[draw=drawColor,line width= 0.4pt,line join=round,line cap=round] (  2.68, 48.88) --
	(  9.37, 51.90) --
	( 16.06, 56.70) --
	( 22.75, 65.91) --
	( 29.44, 76.51) --
	( 36.13, 85.95) --
	( 42.83, 93.89) --
	( 49.52, 95.91) --
	( 56.21,100.44) --
	( 62.90,104.67) --
	( 69.59,111.01);
\definecolor{drawColor}{RGB}{128,177,211}

\path[draw=drawColor,line width= 0.4pt,line join=round,line cap=round] (  2.68, 48.90) --
	(  9.37, 52.79) --
	( 16.06, 58.13) --
	( 22.75, 65.26) --
	( 29.44, 77.11) --
	( 36.13, 85.08) --
	( 42.83, 91.52) --
	( 49.52, 96.74) --
	( 56.21,105.03) --
	( 62.90,105.41) --
	( 69.59,102.91);
\end{scope}
\begin{scope}
\path[clip] (  0.00,  0.00) rectangle ( 72.27,115.63);
\definecolor{drawColor}{RGB}{0,0,0}

\node[text=drawColor,anchor=base,inner sep=0pt, outer sep=0pt, scale=  1.00] at ( 36.13, 13.20) {Prediction};

\node[text=drawColor,anchor=base,inner sep=0pt, outer sep=0pt, scale=  1.00] at ( 36.13,  2.40) {Enron};
\end{scope}
\end{tikzpicture}\hspace{-0.2mm}
\begin{tikzpicture}[x=1pt,y=1pt]
\definecolor{fillColor}{RGB}{255,255,255}
\path[use as bounding box,fill=fillColor,fill opacity=0.00] (0,0) rectangle ( 72.27,115.63);
\begin{scope}
\path[clip] (  0.00,  0.00) rectangle ( 72.27,115.63);
\definecolor{drawColor}{RGB}{0,0,0}

\path[draw=drawColor,line width= 0.4pt,line join=round,line cap=round] (  0.00, 45.60) --
	( 72.27, 45.60) --
	( 72.27,115.63) --
	(  0.00,115.63) --
	(  0.00, 45.60);
\end{scope}
\begin{scope}
\path[clip] (  0.00,  0.00) rectangle ( 72.27,115.63);
\definecolor{drawColor}{RGB}{0,0,0}

\path[draw=drawColor,line width= 0.4pt,line join=round,line cap=round] (  2.68, 45.60) -- ( 69.59, 45.60);

\path[draw=drawColor,line width= 0.4pt,line join=round,line cap=round] (  2.68, 45.60) -- (  2.68, 39.60);

\path[draw=drawColor,line width= 0.4pt,line join=round,line cap=round] (  9.37, 45.60) -- (  9.37, 39.60);

\path[draw=drawColor,line width= 0.4pt,line join=round,line cap=round] ( 16.06, 45.60) -- ( 16.06, 39.60);

\path[draw=drawColor,line width= 0.4pt,line join=round,line cap=round] ( 22.75, 45.60) -- ( 22.75, 39.60);

\path[draw=drawColor,line width= 0.4pt,line join=round,line cap=round] ( 29.44, 45.60) -- ( 29.44, 39.60);

\path[draw=drawColor,line width= 0.4pt,line join=round,line cap=round] ( 36.13, 45.60) -- ( 36.13, 39.60);

\path[draw=drawColor,line width= 0.4pt,line join=round,line cap=round] ( 42.83, 45.60) -- ( 42.83, 39.60);

\path[draw=drawColor,line width= 0.4pt,line join=round,line cap=round] ( 49.52, 45.60) -- ( 49.52, 39.60);

\path[draw=drawColor,line width= 0.4pt,line join=round,line cap=round] ( 56.21, 45.60) -- ( 56.21, 39.60);

\path[draw=drawColor,line width= 0.4pt,line join=round,line cap=round] ( 62.90, 45.60) -- ( 62.90, 39.60);

\path[draw=drawColor,line width= 0.4pt,line join=round,line cap=round] ( 69.59, 45.60) -- ( 69.59, 39.60);

\node[text=drawColor,anchor=base,inner sep=0pt, outer sep=0pt, scale=  1.00] at (  2.68, 26.07) {0};

\node[text=drawColor,anchor=base,inner sep=0pt, outer sep=0pt, scale=  1.00] at ( 22.75, 26.07) {0.3};

\node[text=drawColor,anchor=base,inner sep=0pt, outer sep=0pt, scale=  1.00] at ( 49.52, 26.07) {0.7};

\node[text=drawColor,anchor=base,inner sep=0pt, outer sep=0pt, scale=  1.00] at ( 69.59, 26.07) {1};
\end{scope}
\begin{scope}
\path[clip] (  0.00, 45.60) rectangle ( 72.27,115.63);
\definecolor{drawColor}{RGB}{0,0,0}

\path[draw=drawColor,line width= 0.4pt,dash pattern=on 4pt off 4pt ,line join=round,line cap=round] (  2.68, 48.19) --
	( 69.59,113.04);
\definecolor{drawColor}{RGB}{141,211,199}

\path[draw=drawColor,line width= 0.4pt,line join=round,line cap=round] (  2.68, 48.48) --
	(  9.37, 52.07) --
	( 16.06, 58.89) --
	( 22.75, 66.81) --
	( 29.44, 76.28) --
	( 36.13, 84.95) --
	( 42.83, 92.41) --
	( 49.52,100.25) --
	( 56.21,105.60) --
	( 62.90,109.98) --
	( 69.59,112.38);
\definecolor{drawColor}{RGB}{255,255,179}

\path[draw=drawColor,line width= 0.4pt,line join=round,line cap=round] (  2.68, 48.48) --
	(  9.37, 52.18) --
	( 16.06, 58.99) --
	( 22.75, 67.44) --
	( 29.44, 76.25) --
	( 36.13, 84.84) --
	( 42.83, 93.34) --
	( 49.52,100.72) --
	( 56.21,105.94) --
	( 62.90,110.32) --
	( 69.59,112.27);
\definecolor{drawColor}{RGB}{190,186,218}

\path[draw=drawColor,line width= 0.4pt,line join=round,line cap=round] (  2.68, 48.48) --
	(  9.37, 52.19) --
	( 16.06, 59.02) --
	( 22.75, 67.13) --
	( 29.44, 76.43) --
	( 36.13, 84.60) --
	( 42.83, 93.33) --
	( 49.52, 99.99) --
	( 56.21,105.62) --
	( 62.90,110.08) --
	( 69.59,112.09);
\definecolor{drawColor}{RGB}{251,128,114}

\path[draw=drawColor,line width= 0.4pt,line join=round,line cap=round] (  2.68, 48.48) --
	(  9.37, 52.06) --
	( 16.06, 58.99) --
	( 22.75, 67.12) --
	( 29.44, 76.06) --
	( 36.13, 84.95) --
	( 42.83, 92.76) --
	( 49.52,100.02) --
	( 56.21,105.86) --
	( 62.90,110.10) --
	( 69.59,112.42);
\definecolor{drawColor}{RGB}{128,177,211}

\path[draw=drawColor,line width= 0.4pt,line join=round,line cap=round] (  2.68, 48.47) --
	(  9.37, 52.10) --
	( 16.06, 59.15) --
	( 22.75, 67.23) --
	( 29.44, 76.47) --
	( 36.13, 85.16) --
	( 42.83, 92.61) --
	( 49.52,100.08) --
	( 56.21,105.67) --
	( 62.90,110.21) --
	( 69.59,112.19);
\end{scope}
\begin{scope}
\path[clip] (  0.00,  0.00) rectangle ( 72.27,115.63);
\definecolor{drawColor}{RGB}{0,0,0}

\node[text=drawColor,anchor=base,inner sep=0pt, outer sep=0pt, scale=  1.00] at ( 36.13, 13.20) {Prediction};

\node[text=drawColor,anchor=base,inner sep=0pt, outer sep=0pt, scale=  1.00] at ( 36.13,  2.40) {Mediamill};
\end{scope}
\end{tikzpicture}
\caption{Calibration of the Binary Relevance Classifiers on all Benchmark Datasets. Colors Indicate Different Folds.}
\label{fig:calibration}
\end{figure*}
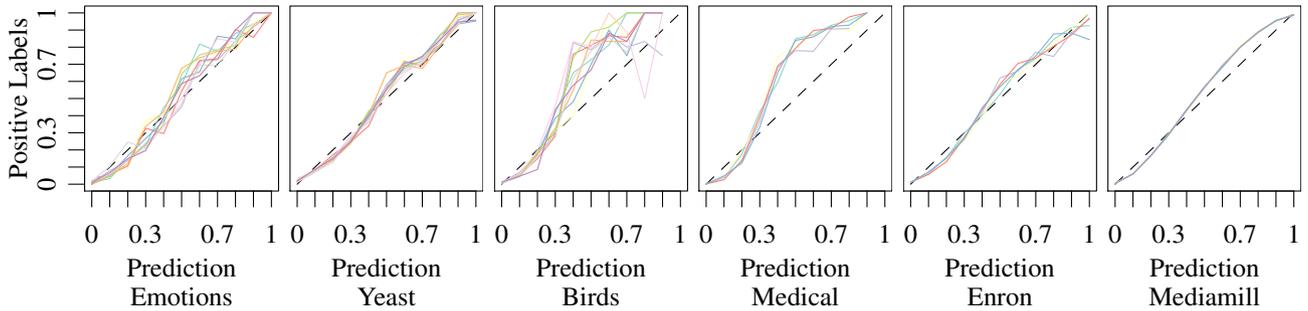

The second hypothesis is that the initial beliefs of the underlying random forests that pRSL relied on were possibly inadequate. This is motivated by the fact that pRSL is the only benchmarked learner that treats the inputs as probabilities, whereas the other learners see them as scores. Additionally, on each of the three aforementioned datasets, BR was outperformed by its competitors in terms of hamming loss by a considerable margin. To investigate this, Figure \ref{fig:calibration} shows the calibration of the random forests' probability estimates, that is the relative amount of positive labels stratified by the probability estimates given on them. The dashed bisector visualizes the ideal case in which the classifier is neither under-- nor overconfident. It can be seen that the random forest is underconfident for birds, medical and slightly for mediamill. To further assess the hypotheses, we calibrated the random forests' probability estimates using Platt scaling \citep{platt2000} and relearned pRSL on the same hyperparameter and initial values settings as before. However, the performance stayed on an equal level despite expected sampling uncertainty. This might indicate that the underconfidence is not a reason but a symptom of the performance. 

As a word of caution, we want to add that the aforementioned hypotheses require more elaborate testing on simulation and real--world datasets. However, such experiments are beyond the scope of this paper.



\section{Conclusion} \label{sec:conclusion}

We introduced pRSL, a learner for multi--label classification that relies on probabilistically extended propositional logic rules. pRSL takes predictions of arbitrary underlying classifiers as input and models the structure between labels by weighting up and down combinations of labels with respect to their contradiction or fulfillment of rules. These rules can be given as any propositional logic formula or learned in a noisy--or form. In the process, we extended the noisy--or gate to multicategorical input. In comparing pRSL to state--of--the--art black--box and interpretable methods on several common benchmark datasets, we found that pRSL is on par with them, albeit it does not outperform them by a larger margin. We ascribe this to pRSL's interpretable nature, and to the fact that it is not yet implemented on GPU, which could allow for a better gradient descent. Moreover, we report promising results of loopy belief propagation algorithms for approximate inference in the marginal and most--probable--explanation case even on larger datasets. 

Focusing on the methodological part in this paper, further advanced classification problems have not been explored. These include learning with missing data, with latent labels, or when dynamically adding and removing classifiers and labels from the ensemble. We will test pRSL's robustness under such conditions in upcoming work, with a focus on the motivating logistics example. Further, we aim to gain more theoretical insight into pRSL's location in the field of similar approaches such as knowledge graphs and probabilistic logic. This might allow to combine pRSL with approaches that model the between--class structures in orthogonal ways.

\begin{contributions} 
    M.~Kirchhof conceived and implemented the idea and wrote the paper.
    L.~Schmid conceived the idea and edited the paper.
    C.~Reining advised on required conditions from a practical perspective.
    M.~t.~Hompel supervised and acquired funding for the project.
    M.~Pauly supervised and edited the paper.
\end{contributions}

\begin{acknowledgements} 
The authors would like to thank Ludger Sandig for fruitful discussions on the Bayesian interpretation of the model. Also, the authors would like to thank Andreas Groll as well as the anonymous reviewers for detailed feedback on earlier drafts of this paper. The work on this publication was supported by Deutsche Forschungsgemeinschaft (DFG) in context of the projects "Transfer Learning for Human Activity Recognition in Logistics" (HO2463/14-2) and "Collaborative Research Center SFB 876 - Providing Information by Resource-Constrained Data Analysis". The authors gratefully acknowledge the computing time provided on the Linux HPC cluster at Technical University Dortmund (LiDO3), partially funded in the course of the Large-Scale Equipment Initiative by the German Research Foundation (DFG) as project 271512359.
\end{acknowledgements}

\bibliography{kirchhofUAI2021}

\clearpage
\newpage

\section*{Supplementary Material}

\subsection{Gradients}

To perform gradient descent, we differentiate the joint log likelihood of the true categories $\boldsymbol{\ell}^* = (\ell_1^*, \dotsc, \ell_J^*)$ by the noisy--or's inhibition probabilities $q^k_{jm}$ for all rules $k = 1, \dotsc, K$, and all $m = 1, \dotsc, M(j)$ categories of labels $j = 1, \dotsc, J$. To simplify notation, we define 
\begin{align*}
\boldsymbol{R}_{-k} :=& (R_1,\dotsc, R_{k-1}, R_{k+1}, \dotsc, R_K),  \\
\boldsymbol{L_{-j}} :=& (L_1, \dotsc, L_{j-1}, L_{j+1} \dotsc, L_J), \\
\boldsymbol{\ell}_{-j} :=& (\ell_1, \dotsc, \ell_{j-1}, \ell_{j+1} \dotsc, \ell_J), \\
\boldsymbol{\ell}_{-j}^* :=& (\ell_1^*, \dotsc, \ell_{j-1}^*, \ell_{j+1}^* \dotsc, \ell_J^*), \\
\mathcal{L}_{j = m} :=& \bigotimes\limits_{v = 1}^{j-1} \{1, \dotsc, M(v)\} \otimes \{m\} \otimes \bigotimes\limits_{v = j+1}^{J} \{1, \dotsc, M(v)\}, \\
\mathcal{L}_{j \neq m} :=& \bigotimes\limits_{v = 1}^{j-1} \{1, \dotsc, M(v)\} \otimes \\
& \{1, \dotsc, m-1, m+1, \dotsc, M(j)\} \otimes \\
& \bigotimes\limits_{v = j+1}^{J} \{1, \dotsc, M(v)\}  \text{ and} \\
q^k_{-j\ell} :=& \prod_{v = 1}^{j-1} q^k_{v\ell_v} \cdot \prod_{v = j+1}^{J} q^k_{v\ell_v} \text{ and} \\
q^k_{-j\ell^*} :=& \prod_{v = 1}^{j-1} q^k_{v\ell_v^*} \cdot \prod_{v = j+1}^{J} q^k_{v\ell_v^*} \,\,.
\end{align*}

\subsubsection{All Labels Known} \label{sec:grAllKnown}

As discussed in the paper, we first differentiate in the case where all true categories $\ell_1^*, \dotsc, \ell_J^*$ are known.

\paragraph{Case 1: Category is incorrect.}

We start for those $q^k_{jm}$ where $m$ is not the true category, that is $\ell_j^* \neq m$:
\begin{align*}
D^k_{jm} :=& \frac{\partial}{\partial q^k_{jm}} P(\boldsymbol{L} = \boldsymbol{\ell}^*|\boldsymbol{R} = \boldsymbol{1}, \boldsymbol{x}) \\
=& \frac{\partial}{\partial q^k_{jm}} \frac{P(R_k = 1, \boldsymbol{L} = \boldsymbol{\ell}^* | \boldsymbol{R}_{-k} = \boldsymbol{1}, \boldsymbol{x})}{P(R_k = 1 | \boldsymbol{R}_{-k} = \boldsymbol{1}, \boldsymbol{x})} \,\,.\\
\end{align*}
From here on, all probabilities stay conditioned on \mbox{$\boldsymbol{R}_{-k}=\boldsymbol{1},\boldsymbol{x}$}, so we will write this condition behind the expression in the following formulas.
\begin{align*}
D^k_{jm} =& \frac{\partial}{\partial q^k_{jm}} \frac{P(R_k = 1, \boldsymbol{L} = \boldsymbol{\ell}^*)}{\sum\limits_{\boldsymbol{\ell} \in \mathcal{L}} (1 - q^k_{j\ell_j} q^k_{-j\ell}) P(\boldsymbol{L} = \boldsymbol{\ell})} \,\,| \,\boldsymbol{R}_{-k} = \boldsymbol{1}, \boldsymbol{x}
\end{align*}
\begin{align*}
=& \frac{\partial}{\partial q^k_{jm}} (P(R_k = 1, \boldsymbol{L} = \boldsymbol{\ell}^*)) \cdot \\ 
& \left( \sum\limits_{\boldsymbol{\ell} \in \mathcal{L}} P(\boldsymbol{L} = \boldsymbol{\ell}) - \sum\limits_{\boldsymbol{\ell} \in \mathcal{L}_{j \neq m}} q^k_{j \ell_j} q^k_{-j \ell} P(\boldsymbol{L} = \boldsymbol{\ell}) - \right. \\
& \left. q^k_{jm} \sum\limits_{\boldsymbol{\ell} \in \mathcal{L}_{j = m}} q^k_{-j\ell} P(\boldsymbol{L} = \boldsymbol{\ell}) \right)^{-1} \,\,| \,\boldsymbol{R}_{-k} = \boldsymbol{1}, \boldsymbol{x} \,\,.
\end{align*}
Simplifying the notation, the above expression can be written as:
\begin{align*}
D^k_{jm} &= \frac{\partial}{\partial q^k_{jm}} \frac{a_1}{a_2 - q^k_{jm} a_3} \,\,| \,\boldsymbol{R}_{-k} = \boldsymbol{1}, \boldsymbol{x} \\
&= \frac{a_1 a_3}{(a_2 - q^k_{jm} a_3)^2} \,\,| \,\boldsymbol{R}_{-k} = \boldsymbol{1}, \boldsymbol{x}
\end{align*}
for suitable choices of $a_1$, $a_2$, and $a_3$. Substituting back and multiplying with $q^k_{jm} (q^k_{jm})^{-1}$ we get:
\begin{align*}
D^k_{jm}=& \frac{P(R_k = 1, \boldsymbol{L} = \boldsymbol{\ell}^*) \sum\limits_{\boldsymbol{\ell} \in \mathcal{L}_{j = m}} q^k_{jm} q^k_{-j\ell} P(\boldsymbol{L} = \boldsymbol{\ell})}{q^k_{jm} P(R_k = 1)^2} \,\,| \,\boldsymbol{R}_{-k} = \boldsymbol{1}, \boldsymbol{x} \\
=& \frac{P(R_k = 1, \boldsymbol{L} = \boldsymbol{\ell}^*) P(L_j = m, R_k = 0)}{q^k_{jm} P(R_k = 1)^2} \,\,| \,\boldsymbol{R}_{-k} = \boldsymbol{1}, \boldsymbol{x} \,\,.
\end{align*}
The first term in the nominator is computationally complex and might get numerically instable with an increasing number of labels $J$, but since we optimize for the log--likelihood, it vanishes via the chain rule:
\begin{align*}
& \frac{\partial}{\partial q^k_{jm}} \log(P(\boldsymbol{L} = \boldsymbol{\ell}^*|\boldsymbol{R} = \boldsymbol{1}, \boldsymbol{x})) \\
=& \frac{1}{P(\boldsymbol{L} = \boldsymbol{\ell}^*|\boldsymbol{R} = \boldsymbol{1}, \boldsymbol{x})} D^k_{jm} \\
=& \frac{P(R_k = 1) P(R_k = 1, \boldsymbol{L} = \boldsymbol{\ell}^*) P(L_j = m, R_k = 0)}{P(\boldsymbol{L} = \boldsymbol{\ell}^*, R_k = 1) q^k_{jm} P(R_k = 1)^2} \,\,| \,\boldsymbol{R}_{-k} = \boldsymbol{1}, \boldsymbol{x} \\
=& \frac{P(L_j = m, R_k = 0)}{q^k_{jm} P(R_k = 1)} \,\,| \boldsymbol{R}_{-k} = \boldsymbol{1}, \boldsymbol{x} \\
=& \frac{P(L_j = m | R_k = 0) (1 - P(R_k = 1))}{q^k_{jm} P(R_k = 1)} \,\,| \,\boldsymbol{R}_{-k} = \boldsymbol{1}, \boldsymbol{x} \,\,.
\end{align*}
Note that $P(L_j = m | R_k = 0)$ is returned for all categories $m = 1, \dotsc, M(j)$ of all labels $L_j, j = 1, \dotsc, J,$ by a single marginal query. Hence, two marginal queries to the network are sufficient to compute the gradient of all inhibition probabilities related to a rule.

\paragraph{Case 2: Category is correct.}

Let us now differentiate for $q^k_{jm}$ where $m$ is the true category, that is $\ell_j^* = m$:
\begin{align*}
D^k_{jm}=& \frac{\partial}{\partial q^k_{jm}} P(\boldsymbol{L} = \boldsymbol{\ell}^*|\boldsymbol{R} = \boldsymbol{1}, \boldsymbol{x}) \\
=& \frac{\partial}{\partial q^k_{jm}} (P(\boldsymbol{L} = \boldsymbol{\ell}^*) - q^k_{j\ell_j^*} q^k_{-j\ell^*} P(\boldsymbol{L} = \boldsymbol{\ell}^*)) \cdot \\
& \left( \sum\limits_{\boldsymbol{\ell} \in \mathcal{L}} P(\boldsymbol{L} = \boldsymbol{\ell}) - \sum\limits_{\boldsymbol{\ell} \in \mathcal{L}_{j \neq m}} q^k_{j\ell_j} q^k_{-j\ell} P(\boldsymbol{L} = \boldsymbol{\ell}) - \right. \\
& \left. q^k_{jm} \sum\limits_{\boldsymbol{\ell} \in \mathcal{L}_{j = m}} q^k_{-j\ell} P(\boldsymbol{L} = \boldsymbol{\ell}) \right)^{-1} \,\,| \boldsymbol{R}_{-k} = \boldsymbol{1}, \boldsymbol{x} \,\,.
\end{align*}
Again, we simplify the notation to make differentiation easier to see:
\begin{align*}
&= \frac{\partial}{\partial q^k_{jm}} \frac{b_1 - q^k_{jm} b_2}{b_3 - q^k_{jm} b_4} \,\,| \,\boldsymbol{R}_{-k} = \boldsymbol{1}, \boldsymbol{x} \\
&= \frac{b_1 b_4 - b_2 b_3}{(b_3 - q^k_{jm} b_4)^2} \,\,| \,\boldsymbol{R}_{-k} = \boldsymbol{1}, \boldsymbol{x} \,\,.
\end{align*}
Substituting back and multiplying by $q^k_{jm} (q^k_{jm})^{-1}$, we get:
\begin{align*}
=& \frac{P(\boldsymbol{L} = \boldsymbol{\ell}^*) (q^k_{jm} b_4 - q^k_{jm} q^k_{-j\ell} b_3)}{q^k_{jm} P(R_k = 1)^2} \,\,| \,\boldsymbol{R}_{-k} = \boldsymbol{1}, \boldsymbol{x} \\
=& (q^k_{jm} P(R_k = 1)^2)^{-1} (P(\boldsymbol{L} = \boldsymbol{l}^*) (P(R_k = 0, L_j = m) - \\
& q^k_{jm} q^k_{-j\ell} (P(L_j = m) + P(L_j \neq m) - \\
&  P(R_k = 0, L_j \neq m)))) \,\,| \,\boldsymbol{R}_{-k} = \boldsymbol{1}, \boldsymbol{x} \\
=& (q^k_{jm} P(R_k = 1)^2)^{-1} (P(\boldsymbol{L} = \boldsymbol{\ell}^*) (P(R_k = 0, L_j = m) - \\
& q^k_{jm} q^k_{-j\ell} (P(L_j = m) + P(R_k = 1, L_j \neq m)))) \,\,| \,\boldsymbol{R}_{-k} = \boldsymbol{1}, \boldsymbol{x} \\
=& (q^k_{jm} P(R_k = 1)^2)^{-1} (P(\boldsymbol{L} = \boldsymbol{\ell}^*) (P(R_k = 0, L_j = m) - \\
& q^k_{jm} q^k_{-j\ell} (P(R_k = 0, L_j = m) + P(R_k = 1, L_j = m) + \\
& P(R_k = 1, L_j \neq m)))) \,\,| \,\boldsymbol{R}_{-k} = \boldsymbol{1}, \boldsymbol{x} \\
=& (q^k_{jm} P(R_k = 1)^2)^{-1} (P(\boldsymbol{L} = \boldsymbol{\ell}^*) (P(R_k = 0, L_j = m) - \\
& q^k_{jm} q^k_{-j\ell} (P(R_k = 0, L_j = m) + P(R_k = 1)))) \,\,| \,\boldsymbol{R}_{-k} = \boldsymbol{1}, \boldsymbol{x} \\
=& (q^k_{jm} P(R_k = 1)^2)^{-1} (P(\boldsymbol{L} = \boldsymbol{l}^*) ((1 - q^k_{jm} q^k_{-j\ell}) \cdot \\ 
& P(R_k = 0, L_j = m) - q^k_{jm} q^k_{-j\ell} P(R_k = 1))) \,\,| \,\boldsymbol{R}_{-k} = \boldsymbol{1}, \boldsymbol{x} \\
=& (q^k_{jm} P(R_k = 1)^2)^{-1} (P(\boldsymbol{L} = \boldsymbol{\ell}^*, R_k = 1) P(R_k = 0, L_j = m) - \\
& P(\boldsymbol{L} = \boldsymbol{\ell}^*) q^k_{jm} q^k_{-j\ell} P(R_k = 1))) \,\,| \,\boldsymbol{R}_{-k} = \boldsymbol{1}, \boldsymbol{x} \,\,.
\end{align*}
Just as before, the computational heavy terms cancel out when we optimize for log--likelihood:
\begin{align*}
& \frac{\partial}{\partial q^k_{jm}} \log(P(\boldsymbol{L} = \boldsymbol{l}^*|\boldsymbol{R} = \boldsymbol{1}, \boldsymbol{x})) \\
=& \frac{P(L_j = m | R_k = 0) (1 - P(R_k = 1))}{q^k_{jm} P(R_k = 1)} - \\ 
& \frac{q^k_{-j\ell^*}}{1 - q^k_{jm} q^k_{-j\ell^*}} \,\,| \,\boldsymbol{R}_{-k} = \boldsymbol{1}, \boldsymbol{x} \,\,.
\end{align*}
This term is similar to the previous gradient and requires the same marginal queries.

\subsubsection{Missing Labels} \label{sec:grUnknowns}

In this case only a subset of labels $\boldsymbol{L}' \subset \boldsymbol{L}$ is known and takes the categories $\boldsymbol{L}' = \boldsymbol{\ell}'$, while the ground truth for all other labels $\boldsymbol{L}^0 = \boldsymbol{L} \backslash \boldsymbol{L}'$ is unknown. In consequence, the optimization goal changes slightly. We define $\mathcal{L}^0, \mathcal{L}^0_{j = m}, \mathcal{L}^0_{j \neq m}$ and $\boldsymbol{\ell}^0$ in analogy to before.

\paragraph{Case 1: Category is known and incorrect.} 

As in the previous section, we will first consider the case where $L_j \in \boldsymbol{L}'$ is known and $m$ is not the true category, that is $\ell_j \neq m$:
\begin{align*}
& \frac{\partial}{\partial q^k_{jm}} P(\boldsymbol{L}' = \boldsymbol{l}'|\boldsymbol{R} = \boldsymbol{1}, \boldsymbol{x}) \\
=& \frac{\partial}{\partial q^k_{jm}} \frac{\sum\limits_{\boldsymbol{\ell}^0 \in \mathcal{L}^0} P(R_k = 1, \boldsymbol{L}' = \boldsymbol{\ell}', \boldsymbol{L}^0 = \boldsymbol{\ell}^0 | \boldsymbol{R}_{-k} = \boldsymbol{1}, \boldsymbol{x})}{P(R_k = 1 | \boldsymbol{R}_{-k} = \boldsymbol{1}, \boldsymbol{x})}
\end{align*}
As the nominator is a factor independent of $q^k_{jm}$, the sum can be placed before the expression and the computations follow those in Section \ref{sec:grAllKnown}. When taking the logarithm, the gradient is divided by $P(\boldsymbol{L}' = \boldsymbol{\ell}'|\boldsymbol{R} = \boldsymbol{1}, \boldsymbol{x})$, so that the resulting gradient remains the same as in Section \ref{sec:grAllKnown}:
\begin{align*}
& \frac{\partial}{\partial q^k_{jm}} \log(P(\boldsymbol{L}' = \boldsymbol{\ell}'|\boldsymbol{R} = \boldsymbol{1}, \boldsymbol{x})) \\
=& \frac{P(L_j = m | R_k = 0) (1 - P(R_k = 1))}{q^k_{jm} P(R_k = 1)} \,\,| \,\boldsymbol{R}_{-k} = \boldsymbol{1}, \boldsymbol{x} \,\,.
\end{align*}

\paragraph{Case 2: Category is known and correct.}

In the case that $m$ is known and the true category, that is $L_j \in \boldsymbol{L}'$ and $\ell_j = m$, the sum can again be factored out. Thus, we can again make use of the gradients from Section \ref{sec:grAllKnown}:
\begin{align*}
& \frac{\partial}{\partial q^k_{jm}} P(\boldsymbol{L}' = \boldsymbol{\ell}'|\boldsymbol{R} = \boldsymbol{1}, \boldsymbol{x}) \\
=& \sum\limits_{\boldsymbol{\ell}^0 \in \mathcal{L}^0} \frac{\partial}{\partial q^k_{jm}} P(\boldsymbol{L}' = \boldsymbol{\ell}', \boldsymbol{L}^0 = \boldsymbol{\ell}^0|\boldsymbol{R} = \boldsymbol{1}, \boldsymbol{x}) \\
=& (q^k_{jm} P(R_k = 1)^2)^{-1} (P(\boldsymbol{L}' = \boldsymbol{\ell}', R_k = 1) P(R_k = 0, L_j = m) - \\
& P(\boldsymbol{L}' = \boldsymbol{\ell}', R_k = 0) P(R_k = 1))) \,\,| \,\boldsymbol{R}_{-k} = \boldsymbol{1}, \boldsymbol{x} \,\,.
\end{align*}
When taking the logarithm, the expression gets slightly more complicated than in Section \ref{sec:grAllKnown}:
\begin{align*}
& \frac{\partial}{\partial q^k_{jm}} \log(P(\boldsymbol{L}' = \boldsymbol{\ell}'|\boldsymbol{R} = \boldsymbol{1}, \boldsymbol{x})) \\
=& (P(\boldsymbol{L}' = \boldsymbol{\ell}'|\boldsymbol{R} = \boldsymbol{1}, \boldsymbol{x}))^{-1} \frac{\partial}{\partial q^k_{jm}} P(\boldsymbol{L}' = \boldsymbol{\ell}'|\boldsymbol{R} = \boldsymbol{1}, \boldsymbol{x}) \\
=& \frac{P(L_j = m | R_k = 0) (1 - P(R_k = 1))}{q^k_{jm} P(R_k = 1)} - \\ 
& \frac{P(\boldsymbol{L}' = \boldsymbol{\ell}', R_k = 0)}{P(\boldsymbol{L}' = \boldsymbol{\ell}', R_k = 1)} \,\,| \,\boldsymbol{R}_{-k} = \boldsymbol{1}, \boldsymbol{x} \\
=& \frac{P(L_j = m | R_k = 0) (1 - P(R_k = 1))}{q^k_{jm} P(R_k = 1)} - \\ 
& \frac{P(R_k = 0 | \boldsymbol{L}' = \boldsymbol{\ell}')}{P(R_k = 1 | \boldsymbol{L}' = \boldsymbol{\ell}')} \,\,| \,\boldsymbol{R}_{-k} = \boldsymbol{1}, \boldsymbol{x} \,\,.
\end{align*}
So, one additional marginal query has to be calculated per rule. Note that this expression simplifies to that of Section \ref{sec:grAllKnown} if all labels are known, that is if $\boldsymbol{L}' = \boldsymbol{L}$, because the conditional probabilities in the second fraction are then the rules conditional probability tables, given by $\boldsymbol{q}$ alone. 

\paragraph{Case 3: Label is unknown.}

Interestingly, the gradients can also be computed for labels that have no ground truth, that is $L_j \in \boldsymbol{L}^0$. Obviously, we do not need to distinguish whether $m$ is correct or not. Let $\boldsymbol{L}^0_{-j} := \boldsymbol{L}^0 \backslash L_j$ and let $q^{k}_{-j\ell} := \prod_{v: L_v \in \boldsymbol{L}'} q^k_{v\ell_v'} \cdot \prod_{v: L_v \in \boldsymbol{L}^0_{-j}} q^k_{v\ell_v}$ for simpler notation.
\begin{align*}
& \frac{\partial}{\partial q^k_{jm}} P(\boldsymbol{L}' = \boldsymbol{\ell}'|\boldsymbol{R} = \boldsymbol{1}, \boldsymbol{x}) \\
=& \frac{\partial}{\partial q^k_{jm}} \frac{\sum\limits_{\boldsymbol{\ell}^0 \in \mathcal{L}^0} (1 - q^k_{j\ell_j^*} q^k_{-j\ell^*}) P(\boldsymbol{L}' = \boldsymbol{\ell}', \boldsymbol{L}^0 = \boldsymbol{\ell}^0)}{P(R_k = 1)} \,\,| \boldsymbol{R}_{-k} = \boldsymbol{1}, \boldsymbol{x}
\end{align*}
\begin{align*}
=& \frac{\partial}{\partial q^k_{jm}} \left(\sum\limits_{\boldsymbol{\ell}^0 \in \mathcal{L}^0} P(\boldsymbol{L}' = \boldsymbol{\ell}', \boldsymbol{L}^0 = \boldsymbol{\ell}^0) - \right. \\
& \left. \sum\limits_{\boldsymbol{\ell}^0 \in \mathcal{L}^0_{-j}} q^k_{j\ell_j^*} q^k_{-j\ell^*} P(\boldsymbol{L}' = \boldsymbol{\ell}', \boldsymbol{L}^0 = \boldsymbol{\ell}^0) - \right. \\
& \left. q^k_{jm} \sum\limits_{\boldsymbol{\ell}^0 \in \mathcal{L}^0_{-j}} q^k_{-j\ell^*} P(\boldsymbol{L}' = \boldsymbol{\ell}', \boldsymbol{L}^0 = \boldsymbol{\ell}^0) \right) \cdot \\
& \left( \sum\limits_{\boldsymbol{\ell} \in \mathcal{L}} P(\boldsymbol{L} = \boldsymbol{\ell}) - \sum\limits_{\boldsymbol{\ell} \in \mathcal{L}_{j \neq m}} q^k_{j\ell_j} q^k_{-j\ell} P(\boldsymbol{L} = \boldsymbol{\ell}) - \right. \\
& \left. q^k_{jm} \sum\limits_{\boldsymbol{\ell} \in \mathcal{L}_{j = m}} q^k_{-j\ell} P(\boldsymbol{L} = \boldsymbol{\ell}) \right)^{-1} \,\,| \boldsymbol{R}_{-k} = \boldsymbol{1}, \boldsymbol{x} \,\,.
\end{align*}
Again, we substitute to make differentiation easier to see:
\begin{align*}
&= \frac{\partial}{\partial q^k_{jm}} \frac{c_1 - q^k_{jm} c_2}{c_3 - q^k_{jm} c_4} \,\,| \,\boldsymbol{R}_{-k} = \boldsymbol{1}, \boldsymbol{x} \\
&= \frac{c_1 c_4 - c_2 c_3}{(c_3 - q^k_{jm} c_4)^2} \,\,| \,\boldsymbol{R}_{-k} = \boldsymbol{1}, \boldsymbol{x} \,\,.
\end{align*}
By substituting back and replacing the sum expressions with corresponding probability terms, we receive:
\begin{align*}
=& (P(R_k = 1)^2)^{-1} (P(\boldsymbol{L}' = \boldsymbol{\ell}') - P(L_j \neq m, \boldsymbol{L}' = \boldsymbol{\ell}', R_k = 0)) \cdot \\
& (q^k_{jm})^{-1} P(L_j = m, R_k = 0) - (q^k_{jm})^{-1} \cdot \\
& P(L_j = m, \boldsymbol{L}' = \boldsymbol{\ell}', R_k = 0) \cdot (P(L_j = m) + \\
& P(L_j \neq m) - P(L_j \neq m, R_k = 0)) \,\,| \,\boldsymbol{R}_{-k} = \boldsymbol{1}, \boldsymbol{x} \\
=& (q^k_{jm} P(R_k = 1)^2)^{-1} (P(L_j = m, \boldsymbol{L}' = \boldsymbol{\ell}') + \\
& P(L_j \neq m, \boldsymbol{L}' = \boldsymbol{\ell}') - P(L_j \neq m, \boldsymbol{L}' = \boldsymbol{\ell}', R_k = 0)) \cdot \\
& P(L_j = m, R_k = 0) - P(L_j = m, \boldsymbol{L}' = \boldsymbol{\ell}', R_k = 0) \cdot \\
& (P(L_j = m) + P(L_j \neq m, R_k = 1))) \,\,| \,\boldsymbol{R}_{-k} = \boldsymbol{1}, \boldsymbol{x} \\
=& (q^k_{jm} P(R_k = 1)^2)^{-1} (P(L_j = m, \boldsymbol{L}' = \boldsymbol{\ell}', R_k = 0) + \\
& P(L_j = m, \boldsymbol{L}' = \boldsymbol{\ell}', R_k = 1) + P(L_j \neq m, \boldsymbol{L}' = \boldsymbol{\ell}', R_k = 1)) \cdot \\
& P(L_j = m, R_k = 0) - P(L_j = m, \boldsymbol{L}' = \boldsymbol{\ell}', R_k = 0) \cdot \\
& (P(L_j = m, R_k = 0) + P(L_j = m, R_k = 1) + \\
& P(L_j \neq m, R_k = 1))) \,\,| \,\boldsymbol{R}_{-k} = \boldsymbol{1}, \boldsymbol{x} \\
=& (q^k_{jm} P(R_k = 1)^2)^{-1} \cdot \\
& (P(L_j = m, \boldsymbol{L}' = \boldsymbol{\ell}', R_k = 1) P(L_j = m, R_k = 0) + \\
& P(L_j = m, \boldsymbol{L}' = \boldsymbol{\ell}', R_k = 0) P(L_j = m, R_k = 0) + \\
& P(L_j \neq m, \boldsymbol{L}' = \boldsymbol{\ell}', R_k = 1) P(L_j = m, R_k = 0) - \\
& P(L_j = m, \boldsymbol{L}' = \boldsymbol{\ell}', R_k = 0) P(L_j = m, R_k = 0) - \\
& P(L_j = m, \boldsymbol{L}' = \boldsymbol{\ell}', R_k = 0) P(L_j = m, R_k = 1) - \\
& P(L_j = m, \boldsymbol{L}' = \boldsymbol{\ell}', R_k = 0) P(L_j \neq m, R_k = 1)) \,\,| \,\boldsymbol{R}_{-k} = \boldsymbol{1}, \boldsymbol{x} \\
=& (q^k_{jm} P(R_k = 1)^2)^{-1} (P(\boldsymbol{L}' = \boldsymbol{\ell}', R_k = 1) P(L_j = m, R_k = 0) - \\
& P(L_j = m, \boldsymbol{L}' = \boldsymbol{\ell}', R_k = 0) P(R_k = 1)) \,\,| \,\boldsymbol{R}_{-k} = \boldsymbol{1}, \boldsymbol{x} \,\,.
\end{align*}
As before, the log likelihood removes the computationally complicated terms:
\begin{align*}
& \frac{\partial}{\partial q^k_{jm}} \log(P(\boldsymbol{L}' = \boldsymbol{\ell}'|\boldsymbol{R} = \boldsymbol{1}, \boldsymbol{x})) \\
=& \frac{P(L_j = m, R_k = 0)}{q^k_{jm} P(R_k = 1)} - \\
& \frac{P(L_j = m, R_k = 0 | \boldsymbol{L}' = \boldsymbol{\ell}')}{q^k_{jm} P(R_k = 1 | \boldsymbol{L}' = \boldsymbol{\ell}')} \,\,| \,\boldsymbol{R}_{-k} = \boldsymbol{1}, \boldsymbol{x} \,\,.
\end{align*}
At this point it can be seen that the gradient reduces to $0$ if no label has a ground truth, that is $L' = \emptyset$, which makes intuitive sense. Applying one last transformation gives a form that is easier to compute:
\begin{align*}
=& \frac{P(L_j = m | R_k = 0) (1 - P(R_k = 1))}{q^k_{jm} P(R_k = 1)} - \\
& \frac{P(L_j = m | R_k = 0, \boldsymbol{L}' = \boldsymbol{\ell}') (1 - P(R_k = 1 | \boldsymbol{L}' = \boldsymbol{\ell}'))}{q^k_{jm} P(R_k = 1 | \boldsymbol{L}' = \boldsymbol{\ell}')} \\ 
& \,\,| \,\boldsymbol{R}_{-k} = \boldsymbol{1}, \boldsymbol{x} \,\,.
\end{align*}
So, overall four marginal queries are required to compute this gradient, or two more than in the case of known labels.

\subsection{Extension of Noisy--Or} \label{sec:extNor}

The ordinary noisy--or gate as defined in \citet{pearl1988} is connected to a set of binary input variables $L_1, \dotsc, L_J$. Each variable can only set the gate $R_k$ to $R_k = 1$ with a probability $1 - q^k_{j1}$ if the variable itself is $L_j = 1$. Thus the conditional probability distribution that defines $R_k$ is:
\begin{align*}
P(R_k = 0 | L_1 = \ell_1, \dotsc, L_J = \ell_J) = \prod\limits_{j = 1}^J (q^k_{j1})^{\ell_j} \,\,.
\end{align*}
In our case, the input variables may have multiple categories $m = 1, \dotsc, M(j)$ and each of these categories can trigger the gate to be $R_k = 1$ with a probability $1 - q^k_{jm}$. To find the conditional probability distribution of $R_k$ in this case, we start by splitting each input variables $L_j$ up into several binary auxiliary variables $L_{jm}$ where
\begin{align*}
P(L_{jm} = 1 | L_j = a) = \begin{cases}
1, a = m \\
0, a \neq m
\end{cases} = \mathbbm{1}_{L_j = m}(L_j) \,\,.
\end{align*}

\begin{figure}
\centering
\begin{tikzpicture}

\node[shape=circle,draw=black] (l1) at (1,0) {$L_1$};
\node[shape=circle] (l3) at (3.5,0.75) {$\ldots$};
\node[shape=circle,draw=black] (lJ) at (6,0) {$L_J$};

\node[shape=circle,draw=black] (l11) at (0,1.5) {$L_{11}$};
\node[shape=circle] (l12) at (1,1.5) {$\ldots$};
\node[shape=circle,draw=black] (l1M) at (2,1.5) {$L_{1M}$};

\node[shape=circle,draw=black] (lJ1) at (5,1.5) {$L_{J1}$};
\node[shape=circle] (lJ2) at (6,1.5) {$\ldots$};
\node[shape=circle,draw=black] (lJM) at (7,1.5) {$L_{JM}$};

\node[shape=circle,draw=black] (r1) at (3.5,3) {$R_k$};

\draw[-{Latex[length=2mm]}] (l1) to (l11);
\draw[-{Latex[length=2mm]}] (l1) to (l1M);
\draw[-{Latex[length=2mm]}] (lJ) to (lJ1);
\draw[-{Latex[length=2mm]}] (lJ) to (lJM);
\draw[-{Latex[length=2mm]}] (l11) to (r1);
\draw[-{Latex[length=2mm]}] (l1M) to (r1);
\draw[-{Latex[length=2mm]}] (lJ1) to (r1);
\draw[-{Latex[length=2mm]}] (lJM) to (r1);
\end{tikzpicture}
\caption{Decomposition of Multicategorical Inputs for a Binary--Input Noisy--Or Gate.}
\label{fig:nor}
\end{figure}
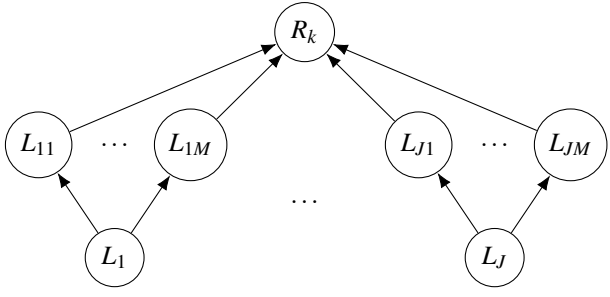

These variables can then be connected to the ordinary binary--input noisy--or gate as visualized in Figure \ref{fig:nor} with their corresponding inhibition probability $q^k_{jm}$. We will now show that this process extends the binary--input noisy--or gate to multicategorical input naturally. We start with the conditional probability via the above described structure with auxiliary variables.

\begin{algorithm}[t]
\centering
\caption{Simulation Dataset Generation}
\label{alg:simData}
\begin{algorithmic}[1]
	\Require nLabels, nRules, nData
	\State pRSL $\gets$ empty model	
	\For{$i$ in $1, \dotsc,$ nLabels}
		\State nCategories $\sim$ $U\{2, \dotsc, 4\}$
		\State Add label node with nCategories to pRSL
	\EndFor	
	\For{$i$ in $1, \dotsc,$ nRules}
		\State nCategories $\sim$ $U\{2, \dotsc, 5\}$
		\State categories $\gets$ Draw nCategories from all 
		\Statex \hspace{2.45cm}$\{1, \dotsc, M(1), \dotsc, 1, \dotsc, M(J)\}$
		\For{each categories}
			\State inhProb $\sim$ Distribution with density 
			\Statex \hspace{2.55cm} $f(x) = 2 \cdot (1 - x) \cdot \mathbbm{1}_{[0, 1]}(x)$
		\EndFor
		\State Add rule with categories and inhProbs to pRSL
	\EndFor
	\For{$i$ in $1, \dotsc,$ nData}
		\For{each label node in pRSL}
			\State classifierOutput[label][i] $\sim$ Dir$(1)$ 
		\EndFor
	\EndFor	
	\State \Return pRSL, classifierOutput
\end{algorithmic}
\end{algorithm}

\begin{align*}
&P(R_k = 0 | \boldsymbol{L} = \boldsymbol{\ell}) \\
=& \sum\limits_{\ell_{11}, \dotsc, \ell_{JM(J)} \ in \{0, 1\}} P(R_k = 0 | L_{11} = \ell_{11}, \dotsc, L_{JM(J)} = \ell_{JM(J)}) \cdot \\
& P(L_{11} = \ell_{11} | L_1 = \ell_1) \cdot \dotsc \cdot P(L_{JM(J)} = \ell_{JM(J)} | L_J = \ell_J) \\
=& \sum\limits_{\ell_{11}, \dotsc, \ell_{JM(J)} \in \{0, 1\}} \prod\limits_{j=1}^J (q^k_{jm})^{L_{jm}} (\mathbbm{1}_{L_1 = 1}(L_1))^{\ell_{11}} \cdot \dotsc \cdot \\
& (\mathbbm{1}_{L_1 = M(1)}(L_1))^{\ell_{1M(1)}} \cdot \dotsc \cdot (\mathbbm{1}_{L_J = 1}(L_J))^{\ell_{J1}} \cdot \dotsc \cdot \\
& (\mathbbm{1}_{L_J = M(J)}(L_J))^{\ell_{JM(J)}}\,\,.
\end{align*}
With $0^0 := 1$, we can see that the only time the term inside the sum is not $0$ is when $L_{jm} = 1 \text{ iff } L_j = m$ for all $j = 1, \dotsc, J$. This leaves open only one possible allocation of the binary auxiliary variables due to their XOR relation within $j$, so that the sum and the auxiliary variables vanish. We finally get a familiar expression that naturally extends the binary noisy--or to the multicategorical case:
\begin{align*}
P(R_k = 0 | \boldsymbol{L} = \boldsymbol{\ell}) = \prod\limits_{j = 1}^J q^k_{j\ell_j} \,\,.
\end{align*}

\subsection{Simulation Dataset Generation} \label{sec:sim}

Algorithm \ref{alg:simData} describes the sampling procedure used to generate the simulation datasets used in Section \ref{sec:approxSim}. In the first ten lines of code, a pRSL model containing nLabels label nodes and nRules rule nodes is randomly generated. In lines $11$ to $13$, the classifier outputs for each classifier node are simulated by dirichlet noise for nData observations. Both the simulated data and the data--generating pRSL model are returned to allow performing approximate marginal and MPE queries on the correct model in Section \ref{sec:approxSim}.

\begingroup
\setlength{\tabcolsep}{4pt} 
\begin{table*}
\caption{Performance of pRSL on Train, Validation, and Test Data. Mean $\pm$ Standard Deviation Between Folds.}
\label{tab:overfitting}
\centering
\begin{tabular}{lcccccc}
\toprule
 & Emotions & Yeast & Birds & Medical & Enron & Mediamill \\
\midrule
\multicolumn{7}{c}{Joint Accuracy (higher = better)} \\
\midrule
Train & $0.351 \pm 0.015$ & $0.227 \pm 0.010$ & $0.516 \pm 0.010$ & $0.500 \pm 0.019$ & $0.153 \pm 0.010$ & $0.146 \pm 0.001$ \\
Validation & $0.339 \pm 0.031$ & $0.251 \pm 0.026$ & $0.516 \pm 0.042$ & $0.497 \pm 0.027$ & $0.154 \pm 0.010$ & $0.149 \pm 0.006$ \\
Test & $0.348 \pm 0.067$ & $0.236 \pm 0.015$ & $0.507 \pm 0.032$ & $0.491 \pm 0.031$ & $0.153 \pm 0.020$ & $0.149 \pm 0.002$\\
\midrule
& \multicolumn{3}{c}{Joint log--Likelihood (higher = better)} & \multicolumn{3}{c}{Label--wise log--Likelihood (higher = better)} \\
\cmidrule(r){1-4} \cmidrule(l){5-7}
Train & $-1.802 \pm 0.077$ & $-3.589 \pm 0.054$ & $-2.534 \pm 0.060$ & $-1.587 \pm 0.022$ & $-6.598 \pm 0.079$ & $-6.585 \pm 0.006$ \\
Validation & $-1.921 \pm 0.241$ & $-3.538 \pm 0.177$ & $-2.532 \pm 0.269$ & $-1.625 \pm 0.051$ & $-6.564 \pm 0.178$ & $-6.563 \pm 0.005$ \\
Test & $-1.839 \pm 0.273$ & $-3.592 \pm 0.085$ & $-2.458 \pm 0.156$ & $-1.565 \pm 0.120$ & $-6.479 \pm 0.242$ & $-6.532 \pm 0.061$\\
\midrule
\multicolumn{7}{c}{Label--wise Hamming Loss (lower = better)} \\
\midrule
Train & $0.182 \pm 0.005$ & $0.191 \pm 0.003$ & $0.043 \pm 0.001$ & $0.015 \pm 0.001$ & $0.046 \pm 0.001$ & $0.027 \pm 0.000$ \\
Validation & $0.181 \pm 0.018$ & $0.188 \pm 0.004$ & $0.042 \pm 0.004$ & $0.015 \pm 0.001$ & $0.046 \pm 0.001$ & $0.027 \pm 0.000$ \\
Test & $0.182 \pm 0.022$ & $0.190 \pm 0.005$ & $0.043 \pm 0.002$ & $0.015 \pm 0.001$ & $0.046 \pm 0.001$ & $0.027 \pm 0.000$\\
\bottomrule
\end{tabular}
\end{table*}
\endgroup

\end{document}